\documentclass[10pt,twocolumn,letterpaper]{article}

\usepackage{cvpr}
\usepackage{times}
\usepackage{epsfig}
\usepackage{graphicx}
\usepackage{amsmath}
\usepackage{amssymb}
\usepackage{color}
\usepackage{epstopdf}
\usepackage{afterpage}

\usepackage[pagebackref=true,breaklinks=true,letterpaper=true,colorlinks,bookmarks=false]{hyperref}

\cvprfinalcopy 

\ifcvprfinal\fi
\begin{document}

\title{Deep Class Aware Denoising}

\author{Tal Remez \textsuperscript{1}\\
{\tt\small talremez@mail.tau.ac.il}
\and
Or Litany \textsuperscript{1}\\
{\tt\small or.litany@gmail.com}
\and
Raja Giryes \textsuperscript{1}\\
{\tt\small raja@tauex.tau.ac.il }
\and
Alex M. Bronstein \textsuperscript{2}\\
{\tt\small bron@cs.technion.ac.il }
\\\textsuperscript{1} School of Electrical Engineering, Tel-Aviv University, Israel
\\\textsuperscript{2} Computer Science Department, Technion - IIT, Israel
}

\maketitle

\begin{abstract}
	
	The increasing demand for high image quality in mobile devices brings forth the need for better computational enhancement techniques, and image denoising in particular. At the same time, the images captured by these devices can be categorized into a small set of semantic classes. However simple, this observation has not been exploited in image denoising until now. In this paper, we demonstrate how the reconstruction quality improves when a denoiser is aware of the type of content in the image. To this end, we first propose a new fully convolutional deep neural network architecture which is simple yet powerful as it achieves state-of-the-art performance even without being class-aware. We further show that a significant boost in performance of up to $0.4$ dB PSNR can be achieved by making our network class-aware, namely, by fine-tuning it for images belonging to a specific semantic class. Relying on the hugely successful existing image classifiers, this research advocates for using a class-aware approach in all image enhancement tasks. 
	
\end{abstract}


\begin{figure*}[t]
	\centering
    \begin{tabular}{c@{\hskip 0.01\textwidth}c@{\hskip 0.01\textwidth}c@{\hskip 0.01\textwidth}c}
		\includegraphics[width = 0.23\textwidth]{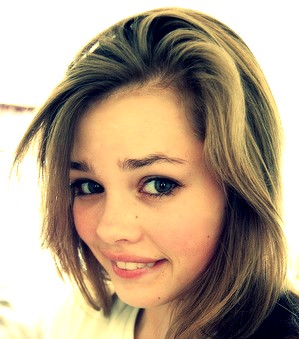} &
		\includegraphics[width = 0.23\textwidth]{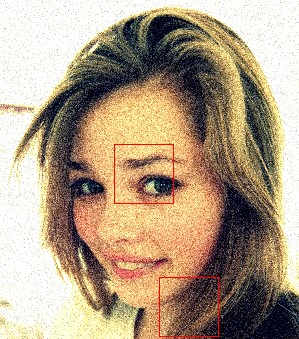} &
        \includegraphics[width = 0.23\textwidth]{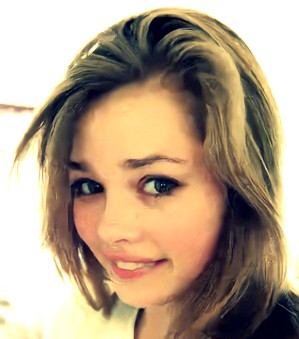} &
		\includegraphics[width = 0.23\textwidth]{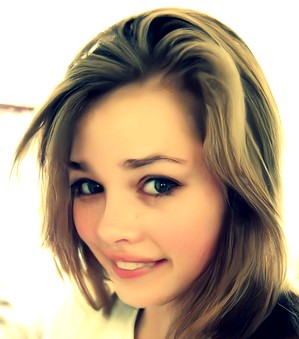} \\
        
	\end{tabular}  
   	\begin{tabular}{c@{\hskip 0.01\textwidth}c@{\hskip 0.01\textwidth}c@{\hskip 0.01\textwidth}c@{\hskip 0.01\textwidth}c@{\hskip 0.01\textwidth}c@{\hskip 0.01\textwidth}c@{\hskip 0.01\textwidth}c}
		\includegraphics[width = 0.11\textwidth]{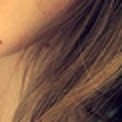} &
        \includegraphics[width = 0.11\textwidth]{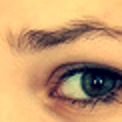} &
		\includegraphics[width = 0.11\textwidth]{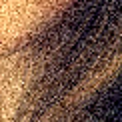} &
        \includegraphics[width = 0.11\textwidth]{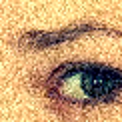} &
   		\includegraphics[width = 0.11\textwidth]{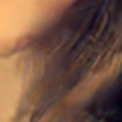} &
        \includegraphics[width = 0.11\textwidth]{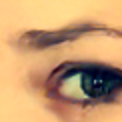} &
   		\includegraphics[width = 0.11\textwidth]{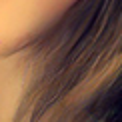} &
        \includegraphics[width = 0.11\textwidth]{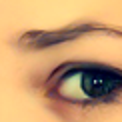} \\  
        
        \multicolumn{2}{c}{Ground truth image}&
        \multicolumn{2}{c}{Noisy image}&
        \multicolumn{2}{c}{Denoised by TNRD \cite{Chen16Trainable}}&
        \multicolumn{2}{c}{Denoised by our method}\\
	\end{tabular}      
    \smallskip 
	\caption{\small \textbf{Perceptual comparison of class-aware and standard denoising.} Our proposed face-specific denoiser produces a visually pleasant result and avoids artifacts commonly introduced by general-purpose denoisers. The reader is encouraged to zoom in for a better view of the artifacts. }
\end{figure*}

\section{Introduction}
The ubiquitous use of mobile phone cameras in the recent decade has set a very high demand on the image quality these devices are expected to produce. On the other hand, the never-ending pursuit of more pixels at smaller form factors puts stringent constraints on the amount of light each pixel is exposed to and results in noisier images. This puts an increasing weight on computational post-processing techniques, in particular on image denoising. 

Many image acquisition artifacts such as low-light noise and camera shake \cite{Delbracio15Burst} can be compensated by image enhancemnet techniques. Denoising in the presense of additive white Gaussian noise is one of the key problems studied in this context. 
While realistic low-light imaging is largely dominated by the Poisson-distributed shot noise, there exist various techniques that allow accurate treatment of non-Gaussian noise sources with a Gaussian denoiser \cite{Makitalo11Optimal, Makitalo14Noise, Rond16Poisson, Sreehari16Plug}.
Moreover, it has been shown in \cite{Chan16Plug, Dar16Postprocessing, Romano16Little, Sreehari16Plug,  Venkatakrishnan13Plug} that having a good Gaussian denoising algorithm allows to solve efficiently  many other image processing problems such as deblurring, inpainting, compression postprocessing and more, without compromising the  reconstruction quality or the need to design a new strategy adapted to a new setting.
In view of these results, it is evident that a good Gaussian denoiser sets the foundation for solving a variety of image reconstruction and enhancement problems. 

Numerous methods have been proposed for removing Gaussian noise from images, including
$k$-SVD \cite{Aharon06KSVD}, non-local means \cite{Buades05Non}, BM3D \cite{dabov2007image} non-local $k$-SVD \cite{Mairal09Non}, field of experts (FoE) \cite{Schmidt10Generative}, Gaussian mixture models (GMM) \cite{Yu12Solving}, non-local Bayes \cite{Lebrun13Nonlocal}, nonlocally centralized sparse representation (NCSR) \cite{Dong13Nonlocally} and simultaneous sparse coding combined with Gaussian scale mixture (SSC-GSM) \cite{Dong15Image}. These techniques have been designed based on some properties of natural images such as the recurrence of patches at different locations or their sparsity in a certain dictionary.

In the past few years, the state-of-the-art in image denoising has been achieved by techniques based on artificial neural networks \cite{burger2012image, Chen16Trainable, vemulapalli2016deep}. 
Neural networks (NNs) are essentially concatenations of basic units (layers), each comprising a linear operation followed by a simple non-linearity, resulting in an intricate highly non-linear response. Currently, they are among the most popular and  powerful tools in machine learning \cite{Bengio09Learning, Deng14Deep, Goodfellow16Deep, LeCun15Deep, Schmidhuber15Deep}.
NN-based approaches have led to state-of-the-art results in numerous tasks in computer vision (e.g. for image classification \cite{He16Deep, Krizhevsky12ImageNet}, video classification \cite{Karpathy14Large}, object detection \cite{Szegedy15Going}, face recognition \cite{Schroff15FaceNet}, and handwriting word recognition \cite{Poznanski16CNN}), speech recognition \cite{Sak15Fast} 
and natural language processing \cite{Bellegarda16State, Hirschberg15Advances, Socher13Recursive, Sutskever14Sequence}, artificial intelligence (e.g., playing videogames \cite{Mnih15Human}
or beating the world Go champion \cite{Silver16Mastering}, which is considered to be a very prominent milestone in the AI community), medical imaging \cite{Greenspan16Guest}, image processing (e.g., image decovolution \cite{Schuler13machine}, inpainting \cite{Pathak16Context} and super-resolution \cite{Bruna16Super, Ledig16Photo, Kim16Accurate}), and more \cite{LeCun15Deep}.

The first neural network to achieve state-of-the-art performance in image denoising has been proposed in \cite{burger2012image}. It is based on a fully connected architecture and therefore requires more training examples at training and much more memory and arithmetic complexity at inference compared to the more recent solution in \cite{vemulapalli2016deep}, which proposes a neural network based on a deep Gaussian Conditional Random Field (DGCRF) model, or the model-based Trainable Nonlinear Reaction Diffusion (TNRD) network introduced in \cite{Chen16Trainable}. 

\subsection*{Contribution}
One of the main elements we find to be missing in the current denosing techniques (and image enhancement strategies in general) is the awareness of the class of images being processed. Such an approach is much needed
as the objects typically photographed by phone camera users belong to a limited number of semantic classes. In this paper, we demonstrate that it is possible to do better image enhancement when the algorithm is \emph{class-aware}. 

We demonstrate this claim on the Gaussian denoising task, for which we propose a novel convolutional neural network (CNN)-based architecture that obtains performance higher than or comparable to the state-of-the-art. The advantage of our architecture is its simple design and the ease of adaptation to new data. 
We fine-tune a pretrained network on several popular image classes and demonstrate a further significant improvement in performance compared to the class-agnostic baseline.

In light of the high performance achieved by modern image classification schemes, the proposed techniqe may be used to improve the image quality in mobile phone camera. To substantiate this claim, we show that the state-of-the-art image classification networks are resilient to the presence of even large amounts of noise.

\section{Class Aware Denoising}
The current theory of patch-based image denoising sets a bound on the achievable performance \cite{Chatterjee10Denoising, levin2011natural, Levin12Patch}. In fact, since existing methods have practically converged to that bound, one may be tempted to deem futile the on-going pursuit of better performance. As it turns out, two possibilities to break this barrier still exist. The first is to use larger patches. This has been proved useful in \cite{burger2012image} where the use of $39 \times 39$ patches allowed to outperform BM3D \cite{dabov2007image} that held the record for many years. A second ``loophole'' which allows a further improvement in denoising performance is to use a better image prior, such as narrowing down the space of images to a more specific class. These two possibilities are not mutually exclusive, and indeed we exploit both. First, as detailed in the sequel, our network has a perceptive field of size $41 \times 41$, which is bigger than the existing practice, while the convolutional architecture keeps the network from becoming prohibitively large. Second, we fine-tune our denoiser to best fit a particular class. The class information can be provided manually by the user, for example when choosing face denoising for cleaning a personal photo collection, or automatically, by applying one of the many existing powerful classification algorithms. 

The idea of combining classification with reconstruction has been previously proposed by 
\cite{baker2002limits} which also dubbed it  \textit{recogstruction}. In their work, 
the authors set a bound on super-resolution performance and showed it can be broken when a face-prior is used. Several other studies have shown that it is beneficial to design a strategy for a specific class. For example, in \cite{Bryt08Compression} it has been shown that the design of a compression algorithm dedicated to faces improves over generic techniques targeting general images. Specifically for the class of faces, several face hallucination methods have been  developed \cite{Wang14Comprehensive}, including face super-resolution and face sketch-photo synthesis techniques. 
In \cite{Joshi10Personal}, the authors showed that given a collection of photos of the same person it is possible to obtain a more faithful reconstruction of the face from a blury image.  In \cite{Iizuka16Let, zhang2016colorful}  class labeling at a pixel-level is used for the colorization of gray-scale images. In \cite{Anwar15Class}, the subspaces attenuated by blur kernels for specific classes are learned, thus improving the deblurring performance. 

Building on the success demonstrated in the aforementioned body of work, in this paper, we propose to use semantic classes as a prior and build class-aware denoisers. Different from previous methods, our model is made class-aware via training and not by design, hence  it may be automatically extended to any type and number of classes. While in this paper we focus on Gaussian denoising, our methodology can be easily extended to much broader class-aware image enhancement, rendering it applicable to many low-level computer vision tasks.


\section{DenoiseNet}
\begin{figure}[tb] \label{fig_denoiseNet}
	\includegraphics[width=0.97\linewidth]{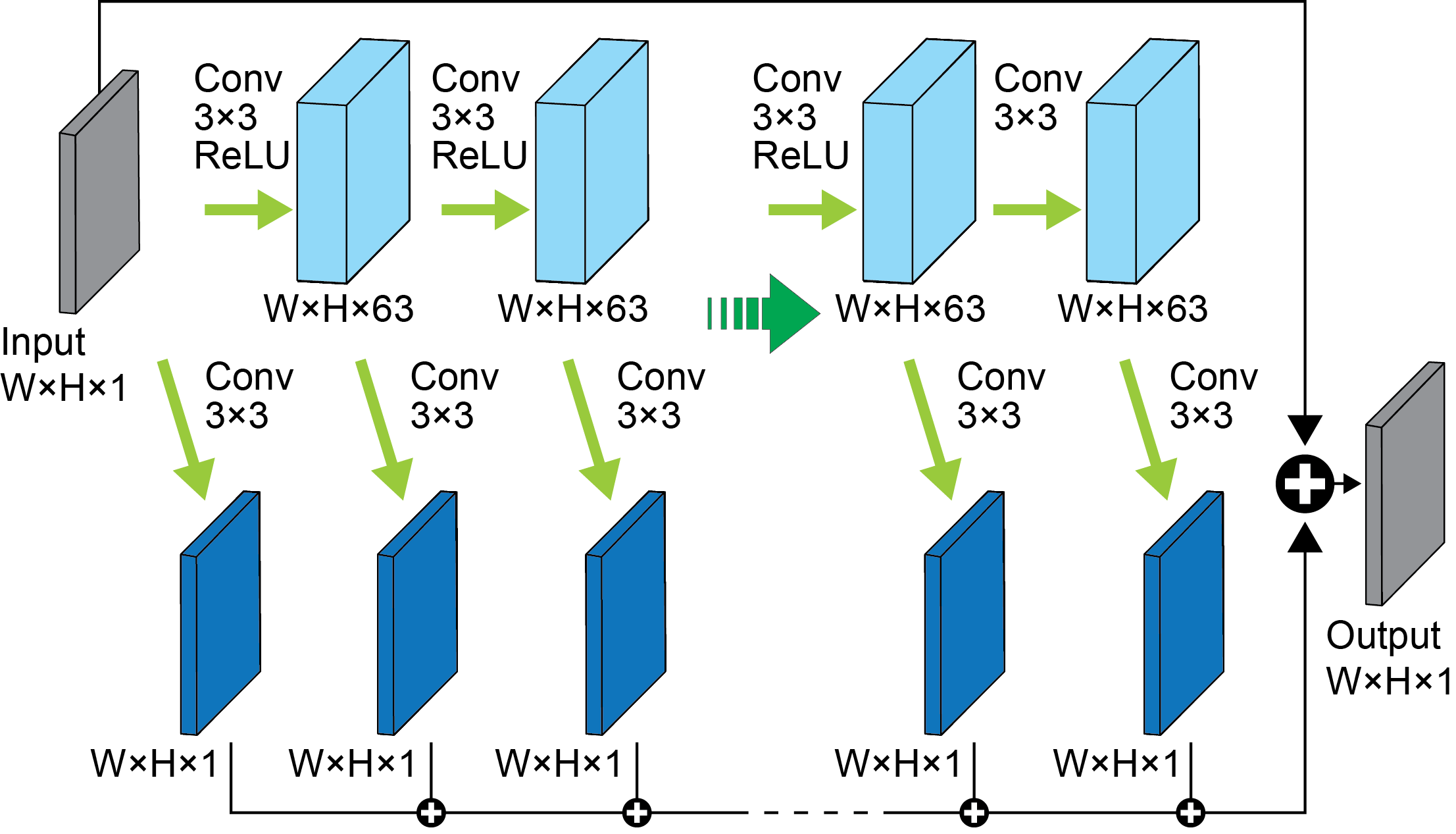}
	\caption{\small \textbf{DenoiseNet fully convolutional architecture.} All convolutions are of size $3 \times 3$ and stride $1$. Convolution resulting feature sizes are listed as $Width \times Height \times \# Channels$. The bottom row of outputs can be viewed as a negative noise components as their sum cancels out the noise.}
\end{figure}

Our network performs additive Gaussian image denoising in a fully convolutional manner. It receives a noisy grayscale image as the input and produces an estimate of the original clean image. The network architecture is shown in Figure \ref{fig_denoiseNet}. The layers at the top row of the diagram calculate features using convolutions of size $3 \times 3$ , stride $1$, and \textit{ReLU} non-linearities. While the layers at the bottom of the diagram can be viewed as negative noise components as their sum cancels out the noise, and are calculated using a single channel convolution of size $3 \times 3$ with stride $1$.
In all experiments we used networks with $20$ layers implemented in TensorFlow \cite{abadi2015tensorflow} and trained it for $160K$ mini-batches on a Titan-X GPU with a set of $8000$ images from the PASCAL VOC dataset \cite{everingham2010pascal}. We used mini-batches of $64$ patches of size $128 \times 128$. Images were converted to YCbCr and the Y channel was used as the input grayscale image after being scaled and shifted to the range of $[-0.5,0.5]$. During training, image patches were randomly cropped and flipped about the vertical axis. To avoid convolution artifacts at the borders of the patches caused by the receptive field of pixels in the deepest layer, we used an $\ell_2$ loss on the central part cropping the outer $21$ pixels during training time and padded the image symmetrically during test time by $21$. Training was done using the ADAM optimizer \cite{DBLP:journals/corr/KingmaB14} with a learning rate of $\alpha=10^{-4}$, $\beta_1=0.9$, $\beta_2=0.999$ and $\epsilon=10^{-8}$. Code and pretrained models will be made available\footnote{\url{https://github.com/TalRemez/deep_class_aware_denoising}}.

\subsection{Simplicity vs capacity}
The choice of network architecture was motivated by the trade off between simplicity and capacity. To best illustrate the concept that class awareness may improve image enhancement algorithms, it was important to incorporate the class via the data, instead of explicitly manipulating  the network architecture. This requires an as-simple-as-possible design. A rather straightforward choice would have been the fully connected architecture proposed by Burger \emph{et al.} \cite{burger2012image}; however, the huge amount of parameters this network uses renders it impractical for many applications.  Alternatively, a very lightweight architecture was proposed by Chen and Pock \cite{Chen16Trainable}; however their model was specifically tailored to their task and, thus, one should be extremely cautious about generalizing any concept demonstrated on it. These two somewhat conflicting paradigms led us to design  a new architecture which is both relatively light-weight while extremely simple to understand and implement. In terms of capacity, we have two orders of magnitude less parameters than the NN proposed by Burger \cite{burger2012image}, but only one order of magnitude more than that introduced by Chen and Pock \cite{Chen16Trainable}. Note that the reduction in the number of parameters does not decrease the receptive field as our model is much deeper. 

%
%

\section{Classification in the presence of noise}
The tacit assumption of our class-aware approach is the ability to determine the class of the noisy input image.
While the goal of this research is not to improve image classification, we argue that the performance of modern CNN based classification algorithm such as \textit{Inception} \cite{Szegedy15Going, DBLP:journals/corr/SzegedyVISW15} or \textit{resNet} \cite{He16Deep} is relatively resilient to a moderate amount of noise. In addition, since we are interested in canonical semantic classes such as \textit{faces} and \textit{pets} which are far coarser than the $1000$ ImageNet classes \cite{ImageNet15}, the task becomes even easier: confusing two breeds of cats is not considered an error. 

\begin{figure}[t] \label{fig_noise_resilience}
	\includegraphics[width=0.45\textwidth]{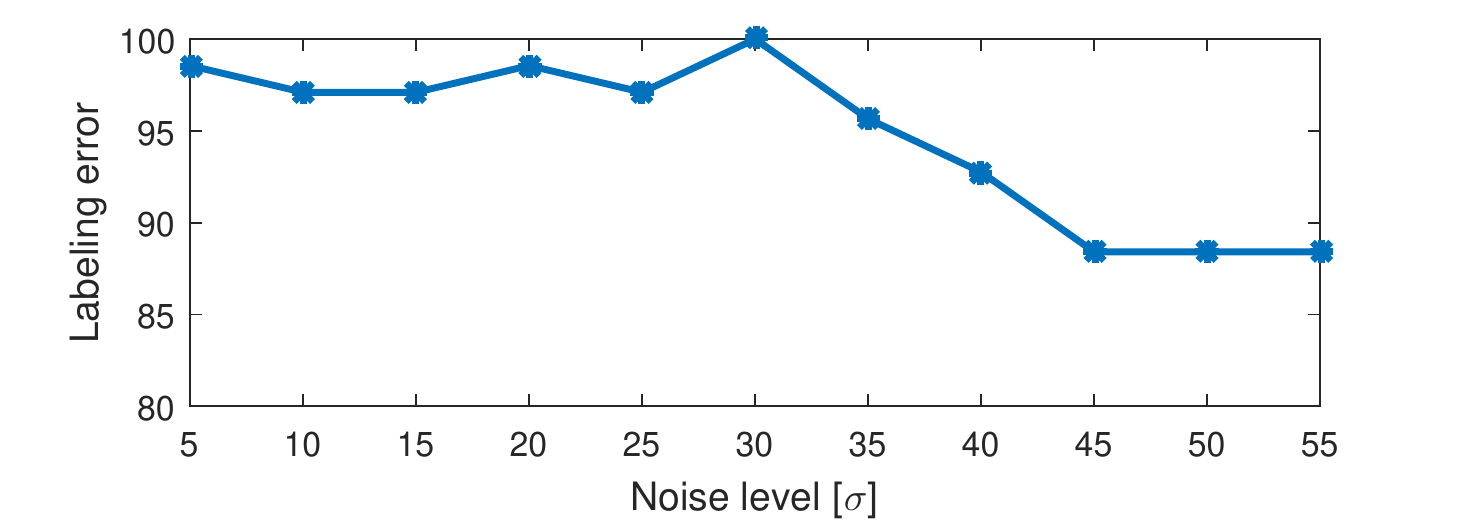}
	\caption{\textbf{Noise resilience of image classification.} The percentage of images on a pre-trained \textit{inception-v3} classifier remains stable exceeds $85\%$ even in the presence of large amount of noise.} \label{fig:resilience}
\end{figure}

The aforementioned networks can be further fine-tuned using noisy examples to increase their resilience to noise. Alternatively, one could simply run a class-agnostic denoiser on the image before plugging it into the classification network. To illustrate the noise resilience property we ran the pre-trained \textit{Inception-v3} \cite{DBLP:journals/corr/SzegedyVISW15} network on a few tens of images from the \emph{pets} class. We then gradually added noise to these images and counted the number of images on which the classifier changed its most confident class to a different class, as visualized in Figure \ref{fig:resilience}. Observe that the network classification remains stable even in the presence of large amount of noise. 

\section{Experiments}
In all experiments in this section our network was trained on $8000$ images from the PASCAL VOC \cite{everingham2010pascal} dataset and was compared to BM3D \cite{dabov2007image}, multilayer perceptrons (MLP) \cite{burger2012image} and TNRD \cite{Chen16Trainable} on the following three test sets: (i) images from PASCAL VOC \cite{pascal-voc-2010}; (ii) a denoising dataset with quantized images from \cite{vemulapalli2016deep}; and (iii)
$68$ test images chosen by \cite{roth2009fields} from the Berkeley segmentation dataset \cite{MartinFTM01}.

\subsection{Class-agnostic denoising}

\paragraph{PASCAL VOC.} In this experiment we tested the denoising algorithms on $1000$ test images from the PASCAL VOC dataset \cite{pascal-voc-2010}. We believe this large and diverse set of images is representative enough to make conclusions about the denoising performance. Table \ref{tab_pascal_results} summarizes performance in terms of average PSNR for all test images contaminated by white Gaussian noise with $\sigma=10$ and up to $\sigma=75$. It is evident that our method outperforms all other methods for both noise levels by over $0.2$ dB.

\begin{table}[h!]
	\centering
	\small
	\begin{tabular}{ l@{\hskip 0.01\textwidth}c@{\hskip 0.01\textwidth}c@{\hskip 0.01\textwidth}c@{\hskip 0.01\textwidth}c@{\hskip 0.01\textwidth}c@{\hskip 0.01\textwidth}c@{\hskip 0.01\textwidth}c  }
		\hline\hline
		$\sigma$					& $10$ & $15$ & $25$ & $35$ & $50$ & $65$ & $75$ \\ \hline
		BM3D						& $34.26$ & $32.10$ & $29.62$ & $28.14$ & $26.61$ & $25.64$ & $25.12$ \\  
		MLP \cite{burger2012image} 	& $34.29$ & $-$ & $29.95$ & $28.49$ & $26.98$ & $26.07$ & $25.54$ \\  
		TNRD \cite{Chen16Trainable}	& $-$ & $32.35$ & $29.90$ & $-$ & $26.91$ & $-$ & $-$ \\  
		DenoiseNet 					& $\textbf{34.87}$ & $\textbf{32.79}$ & $\textbf{30.36}$ & $\textbf{28.88}$ & $\textbf{27.32}$ & $\textbf{26.30}$ & $\textbf{25.74}$	\\ \hline \hline
	\end{tabular}  
	\vspace{2mm}
	\caption{\small \textbf{Performance on PASCAL VOC.} Average PSNR values on a $1000$ image test set. Our method outperforms all other methods for all noise levels.}
	\label{tab_pascal_results}	
\end{table}

\begin{figure}[t] 
	\includegraphics[width=0.48\textwidth]{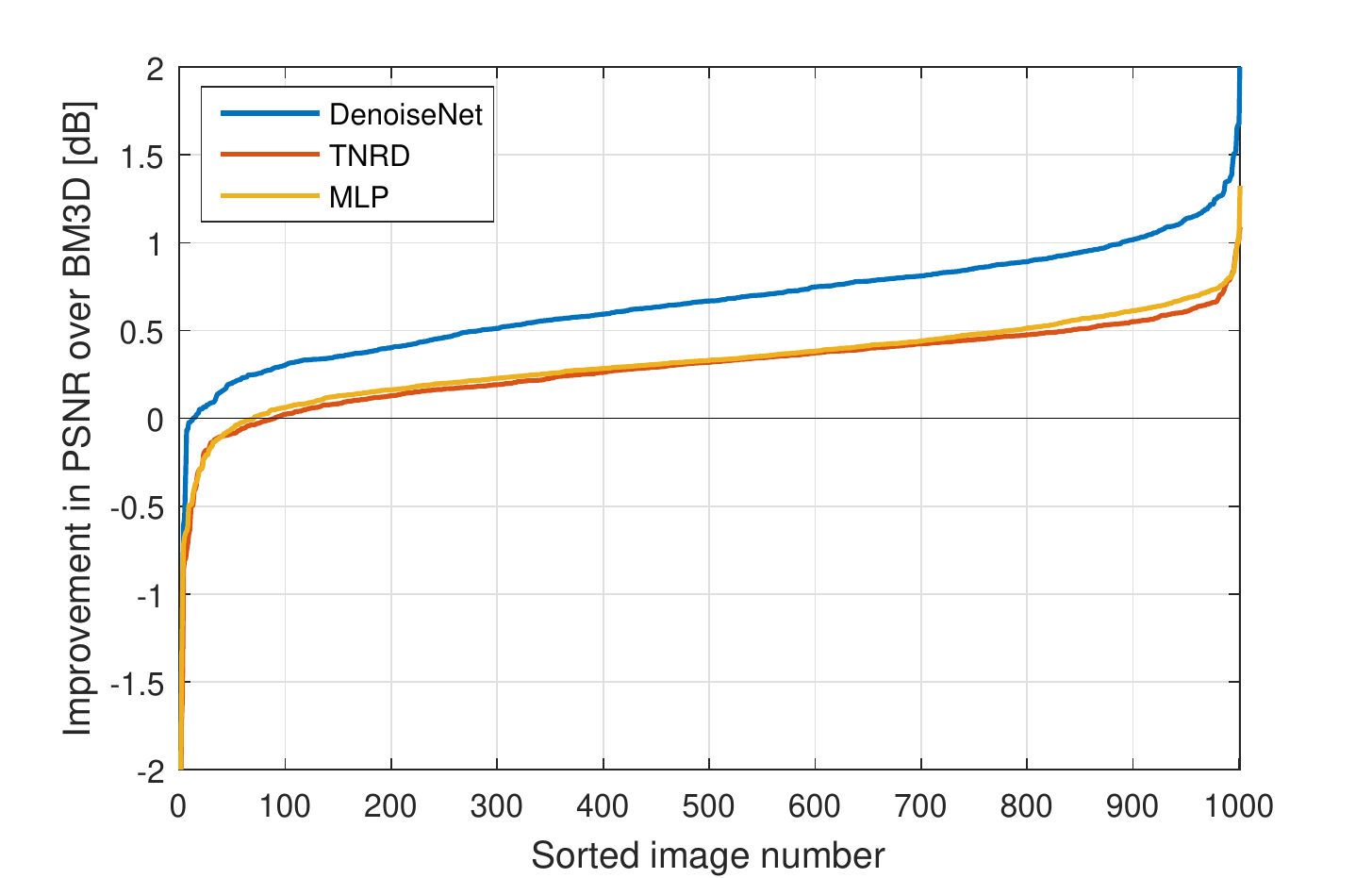}
	\caption{\textbf{Comparison of performance profile relative to BM3D.}  Image indices are sorted in ascending order of performance gain relative to BM3D. The improvement of our method over two competing algorithms is demonstrated by (i) a noticeable decrease of the zero-crossing point, and (ii) consistently higher values of gain over BM3D. The distribution reveals the statistical significance of the reported improvement. The comparison was made on images from PASCAL VOC.} 
	\label{pascal_s_curve}
\end{figure}

\begin{figure}[t] 
	\centering
	\includegraphics[width=0.3\textwidth]{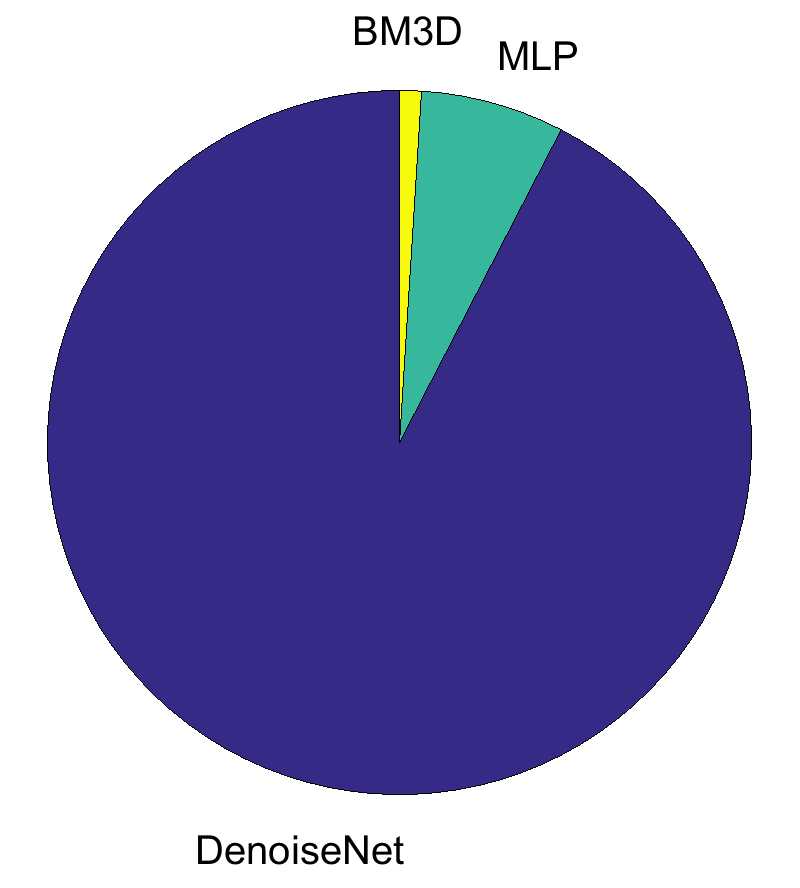}
	\caption{\textbf{Top performance distribution on PASCAL VOC test set.} Percentage of images on which a denoising algorithm performed the best. Our method wins on $92.4\%$ of the images, whereas MLP, BM3D, and TNRD win on $6.6\%, 1\%$ and $0\%$ respectively.} 
	\label{pascal_pie}
\end{figure}

To examine the statistical significance of the improvement our method achieves, in Figure \ref{pascal_s_curve} we compare the gain in performance with respect to BM3D achieved by our method, MLP and TNRD. Image indices are sorted in ascending order of performance gain. A smaller zero-crossing value affirms our method outperforms BM3D on a larger portion of the dataset than the competitors. The plot visualizes the large and consistent improvement in PSNR achieved by DenoiseNet.
A summary of the number of images on which each algorithm performed the best is presented in Figure \ref{pascal_pie}.

\paragraph{Berkeley segmentation dataset.}
In this experiment we tested the performance of our method, trained on PASCAL VOC, on the a set of $68$ images selected by \cite{roth2009fields} from Berkeley segmentation dataset \cite{MartinFTM01}. Even though these test images belong to a different dataset, Figure \ref{tab_bsds_nonquantized_results} shows that our method outperforms previous methods for all sigma values.
\begin{table}[h!]
	\centering
	\small
	\begin{tabular}{ l@{\hskip 0.01\textwidth}c@{\hskip 0.01\textwidth}c@{\hskip 0.01\textwidth}c@{\hskip 0.01\textwidth}c@{\hskip 0.01\textwidth}c@{\hskip 0.01\textwidth}c@{\hskip 0.01\textwidth}c  }
		\hline\hline
		$\sigma$					& $10$ & $15$ & $25$ & $35$ & $50$ & $65$ & $75$ \\ \hline
		BM3D						& $33.31$ & $31.10$ & $28.57$ & $27.08$ & $25.62$ & $24.68$ & $24.20$ \\  
		MLP \cite{burger2012image} 	& $33.50$ & $-$ & $28.97$ & $27.48$ & $26.02$ & $25.10$ & $24.58$ \\  
		TNRD \cite{Chen16Trainable}	& $-$ & $31.41$ & $28.91$ & $-$ & $25.95$ & $-$ & $-$ \\  
		DenoiseNet 					& $\textbf{33.58}$ & $\textbf{31.44}$ & $\textbf{29.04}$ & $\textbf{27.56}$ & $\textbf{26.06}$ & $\textbf{25.12}$ & $\textbf{24.61}$	\\ \hline \hline
	\end{tabular}   
	\vspace{2mm}
	\caption{\small \textbf{Performance on images from Berkeley segmentation dataset.} Average PSNR values on a test set of $68$ images selected by \cite{roth2009fields}. Our method outperforms all others for all noise levels.
		\vspace{-4mm}}
	\label{tab_bsds_nonquantized_results}	
\end{table}


\paragraph{Quantized noise.}
Even though our network has not been explicitly trained to treat quantized noisy images, we evaluated its performance on $300$ such images from \cite{vemulapalli2016deep}. Results are reported in Table \ref{tab_public_denoise_results}. The set contains $100$ test images from the Berkeley segmentation dataset and additional $200$ images from the PASCAL VOC 2012 \cite{pascal-voc-2012} dataset. All images have been quantized to $8$ bits in the range $[0,255]$. For the noise level of $\sigma=25$ our methods outperforms previous methods but fails to do so for $\sigma=50$.
\begin{table}[h]
	\centering
	\begin{tabular}{ l c  c  }
		\hline \hline
		& $\sigma=25$		& $\sigma=50$  		\\ \hline
		BM3D						& $28.21$ 			& $24.43$ 			\\  
		MLP \cite{burger2012image} 	& $28.58$ 			& $\textbf{25.20}$	\\  
		TNRD \cite{Chen16Trainable}	& $28.46$			& $24.57$			\\ 
		DenoiseNet 					& $\textbf{28.71}$	& $24.75$ 			\\ \hline\hline
	\end{tabular}  
	\vspace{2mm}
	\caption{\small \textbf{Performance on quantized test images from \cite{vemulapalli2016deep}.} Images have been clipped to a range of $[0,255]$ and quantized to $8$ bits. PSNR values for two different noise levels are reported.  
	} 
	\label{tab_public_denoise_results}	
\end{table}


\subsection{Class-aware denoising}
\begin{figure}[!htb] 
	\includegraphics[width=0.47\textwidth]{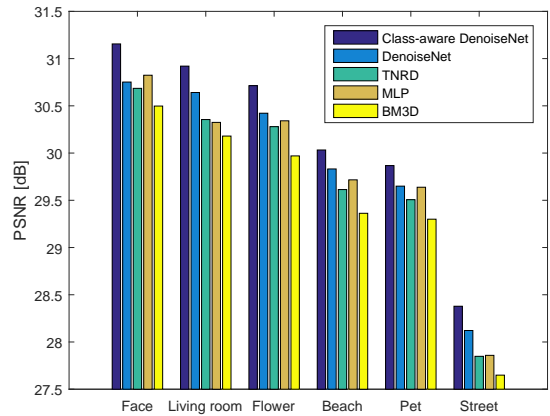}
	\caption{\textbf{Class-aware denoising performance on ImageNet.} Average PSNR values for different methods on images belonging to six different semantic classes. It is evident that the class-specific fine-tuned models outperform all other methods. In addition being class-aware enables to gain up to $0.4$ dB PSNR copared to our class-agnostic network.}
	\label{fig_class_bar}
\end{figure}

This experiment evaluates the boost in performance resulting from fine-tunning a denoiser on a set of images belonging to a particular class. In order to do so we collected images from ImageNet \cite{ImageNet15} of the following six classes: \emph{face}, \emph{pet}, \emph{flower}, \emph{beach}, \emph{living room}, and \emph{street}. The $1,500$ images per class were split into train ($60\%$), validation ($20\%$) and test ($20\%$) sets. We then trained a separate class-aware denoiser for each of the six classes. This was done by fine-tuning our class-agnostic model, that had been trained on PASCAL VOC, using the images from ImageNet.
The performance of the class-aware denoisers was compared to its class-agnostic counterpart as well as to other denoising methods. Average PSNR values summarized in Figure \ref{fig_class_bar} demonstrate that our class-aware models outperforms our class-agnostic network, BM3D \cite{dabov2007image}, multilayer perceptrons (MLP) \cite{burger2012image} and TNRD \cite{Chen16Trainable}. Notice how 
class-awareness boosts performance by up to $0.4dB$.

\subsection{Cross-class denoising}
\begin{figure}[]
	\centering
	\setlength{\tabcolsep}{0.2em}
	\begin{tabular}{ c c c c }
		
		\footnotesize{Ground truth}& \footnotesize{Noisy image} & \footnotesize{Correct denoiser} & \footnotesize{Wrong denoiser}\smallskip \\ 
		
		&   &  \footnotesize{\textit{face}-specific} & \footnotesize{\textit{street}-specific}  \\
		\includegraphics[width = 0.11\textwidth]{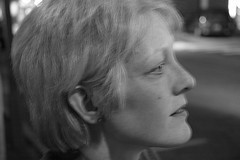} &
		\includegraphics[width = 0.11\textwidth]{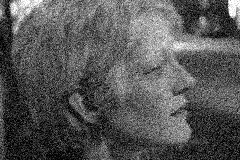} &
		\includegraphics[width = 0.11\textwidth]{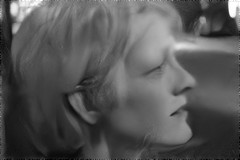} &
		\includegraphics[width = 0.11\textwidth]{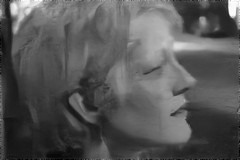} \smallskip\\  
		
		&   & \footnotesize{\textit{pet}-specific} & \footnotesize{\textit{living room}-specific} \\
		\includegraphics[width = 0.11\textwidth]{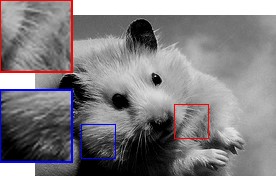} &
		\includegraphics[width = 0.11\textwidth]{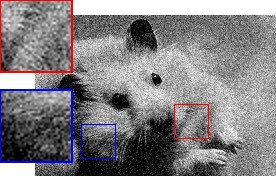} &
		\includegraphics[width = 0.11\textwidth]{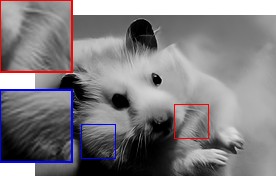} &
		\includegraphics[width = 0.11\textwidth]{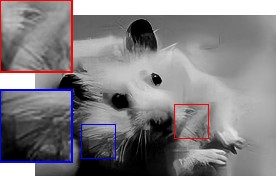} \smallskip\\  

		&   & \footnotesize{\textit{street}-specific} & \footnotesize{\textit{face}-specific} \\    
		\includegraphics[width = 0.11\textwidth]{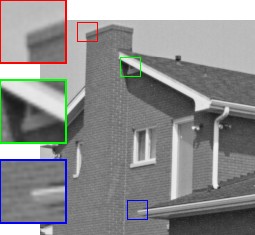} &
		\includegraphics[width = 0.11\textwidth]{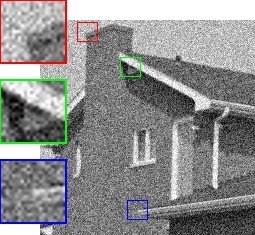} &
		\includegraphics[width = 0.11\textwidth]{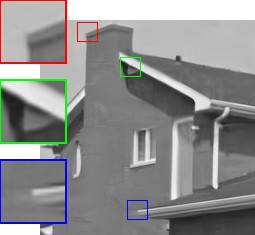} &
		\includegraphics[width = 0.11\textwidth]{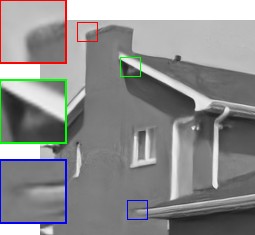} \smallskip\\          

		&   & \footnotesize{\textit{face}-specific} & \footnotesize{\textit{street}-specific} \\
		\includegraphics[width = 0.11\textwidth]{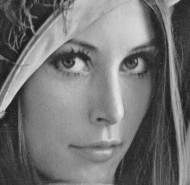} &
		\includegraphics[width = 0.11\textwidth]{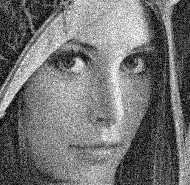} &
		\includegraphics[width = 0.11\textwidth]{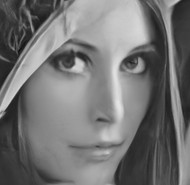} &                 					\includegraphics[width = 0.11\textwidth]{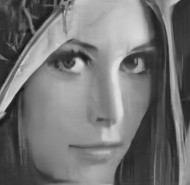} \smallskip\\  

	\end{tabular}   \\
	\caption{\small \textbf{Cross-class denoising.} 
		Representative outputs of DenoiseNet denoisers fine-tuned to the class of the inpuit image (third column from left), and to a mismatched class (rightmost column). The reader is encouraged to zoom in for a better view of the artifacts.}
	\label{fig_using_wrong_denoiser}
\end{figure}

To further demonstrate the effect of refining a denoiser to a particular class, we tested each class-specific denoiser on images belonging to other classes. The outcome of this mismatch is evident both qualitatively and quantitatively. The top row of Figure \ref{fig_using_wrong_denoiser} presents a comparison of class-aware denoisers fine-tuned to the \textit{street} and \textit{face} image classes applied to a noisy image of a face. The denoiser tuned to the street class produces noticeable artifacts around the eye, cheek and hair areas. Moreover, the edges appear too sharp and seem to favor horizontal and vertical edges. This is not very surprising as street images contain mainly man-made rectangle shaped structures. In the second row, strong artifacts appear on the hamster's fur when the image is processed by \textit{living room}-specific denoiser. The \textit{pet}-specific denoiser, on the other hand, produces a much more naturally looking result. Additional examples demonstrating artifacts caused by the mismatch on the canonical images \textit{House} and \textit{Lena} are presented in the bottom two rows. Notice how the \textit{street}-specific denoiser reconstructs sharp boundaries of the building whereas the \textit{face}-specific counterpart smears them. 

\begin{figure}[] 
	\includegraphics[width=0.47\textwidth]{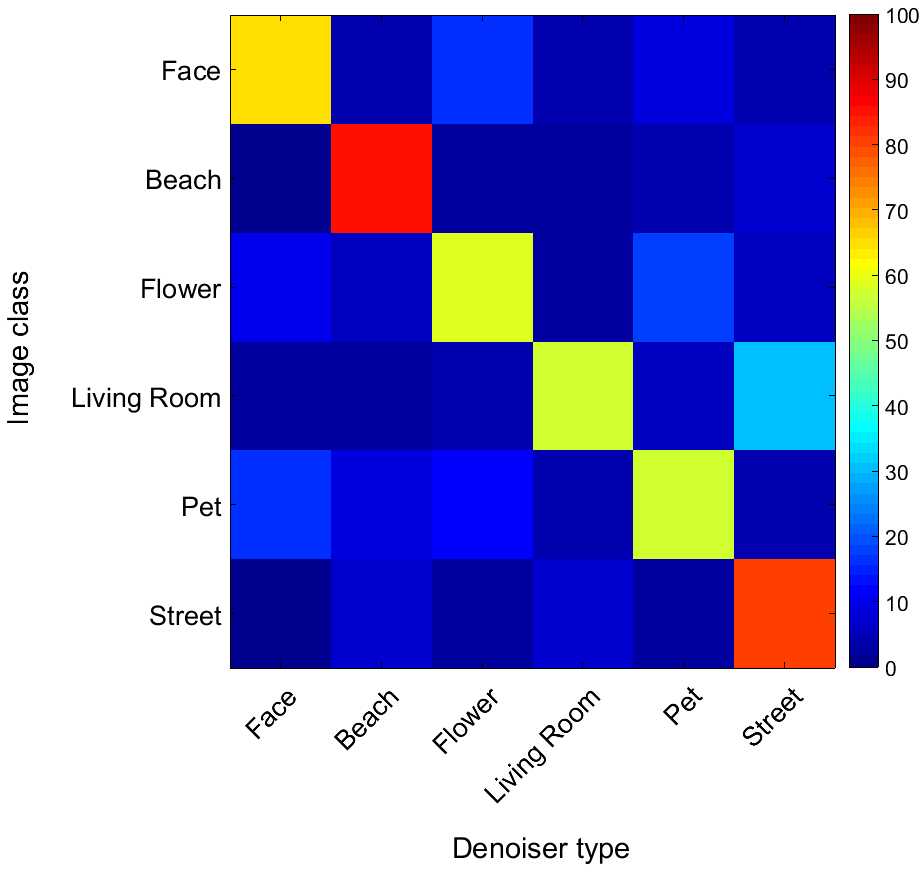}
	\caption{\textbf{Denoiser performance per semantic class.} Each row represents a specific semantic class of images while class-aware denoisers are represented as columns. The $(i,j)$-th element in the confusion matrix shows the probability of the $j$-th class-aware denoiser to outperform all other denoisers on the $i$-th class of images.}
	\label{fig_confusion}
\end{figure}

To quantify the effect of mismatching we evaluated the percentage of wins of every fine-tuned denoiser on each type of image class. A win means that a particular denoiser produced the highest PSNR among all the others. A confusion matrix for all combinations of class-specific denoisers and image classes is presented in Figure \ref{fig_confusion}.
We conclude that applying a denoiser of the same class as the image results in the best performance.

\begin{figure*}[]
	\centering	
	\begin{tabular}{ c@{\hskip 0.008\textwidth}c@{\hskip 0.008\textwidth}c@{\hskip 0.008\textwidth}c@{\hskip 0.008\textwidth}c}
		\includegraphics[width = 0.19\textwidth]{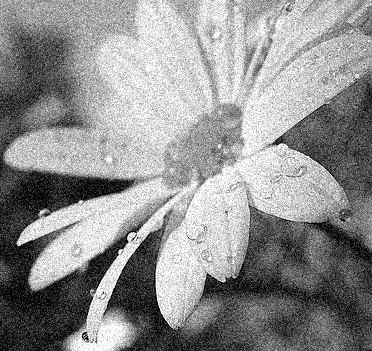} &
		\includegraphics[width = 0.19\textwidth]{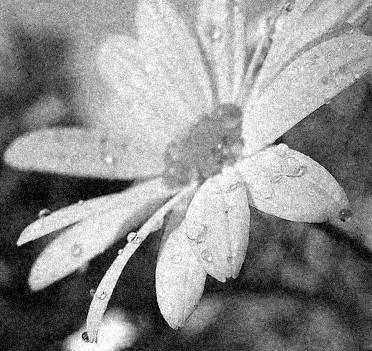} &
		\includegraphics[width = 0.19\textwidth]{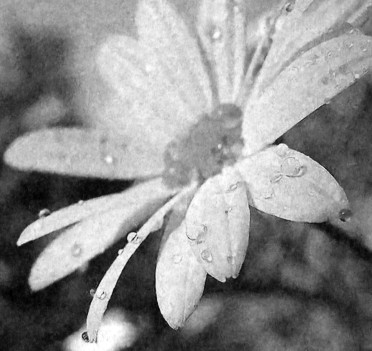} &
		\includegraphics[width = 0.19\textwidth]{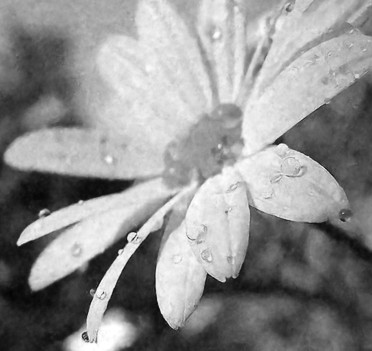} &
		\includegraphics[width = 0.19\textwidth]{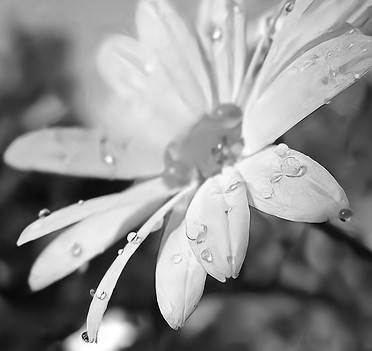} \\  
		Noisy input &   &  &   & Output \\
		
		\includegraphics[width = 0.19\textwidth]{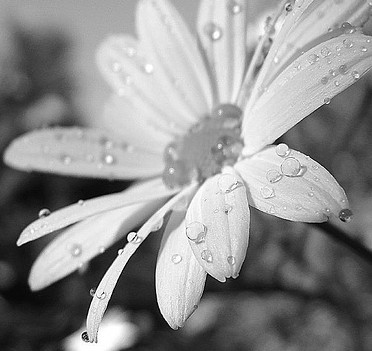} &
		\includegraphics[width = 0.19\textwidth]{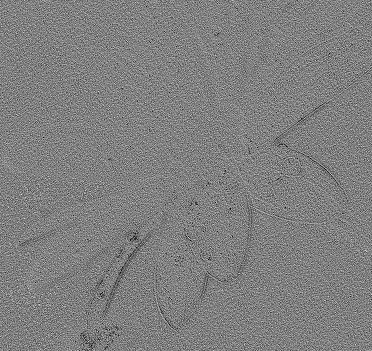} &
		\includegraphics[width = 0.19\textwidth]{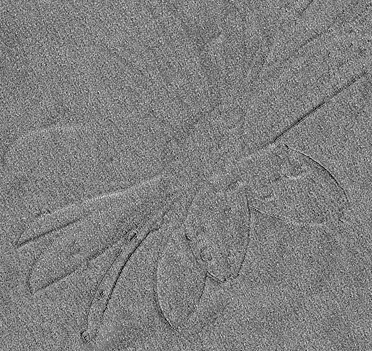} &
		\includegraphics[width = 0.19\textwidth]{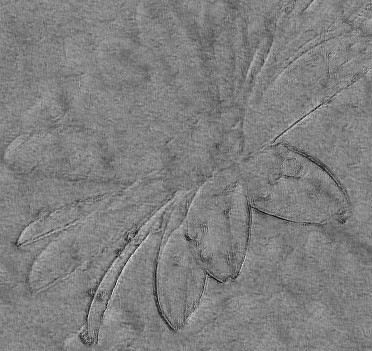} &
		\includegraphics[width = 0.19\textwidth]{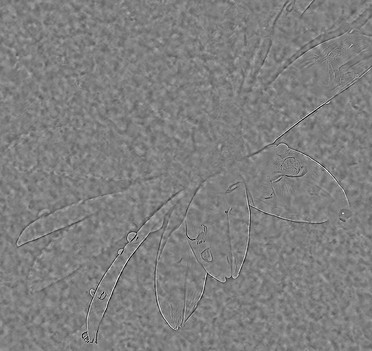} \\    
		
		Ground truth & Layer 5 & Layer 10 & Layer 15 & Layer 20 \\
		
	\end{tabular}   \\
	\smallskip
	\caption{\small \textbf{Gradual denoising process by \textit{flower}-specific DenoiseNet.} The top row presents the noisy image (left) and the intermediate result obtained by removing the noise estimated up to the respective layer depth. 
		The second row presents the ground truth image (left) and the noise estimates produced by individual layers; the noise images have been scaled for display purposes. We encourage the reader to zoom-in onto the images to best view the fine details and noise.}
	\label{fig_denoise_flow}
\end{figure*}

\subsection{Network noise estimation}
\label{sec_noise_estimation}
\begin{figure*}[]
	\begin{centering}
		\begin{tabular}{c@{\hskip 0.007\textwidth}c@{\hskip 0.005\textwidth}c@{\hskip 0.005\textwidth}c@{\hskip 0.005\textwidth}c@{\hskip 0.005\textwidth}c@{\hskip 0.005\textwidth}c}
			\hspace{-1mm}\parbox[b][4em][s]{0.16\textwidth}{Ground truth}&
			\includegraphics[height = 0.15\textwidth]{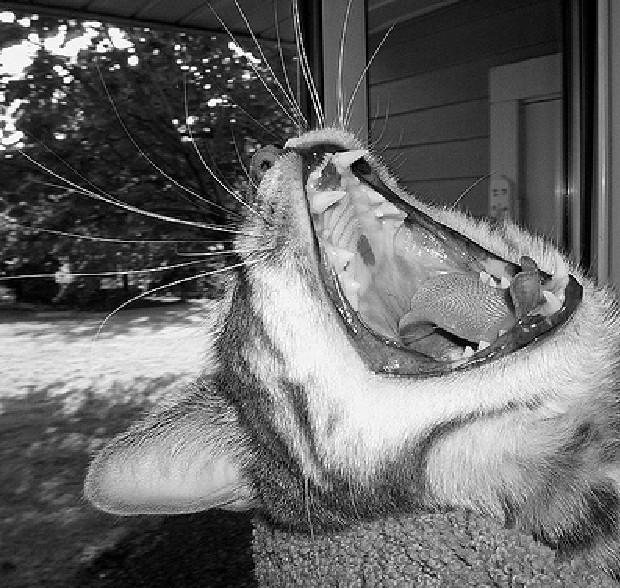} &
			\includegraphics[height = 0.15\textwidth]{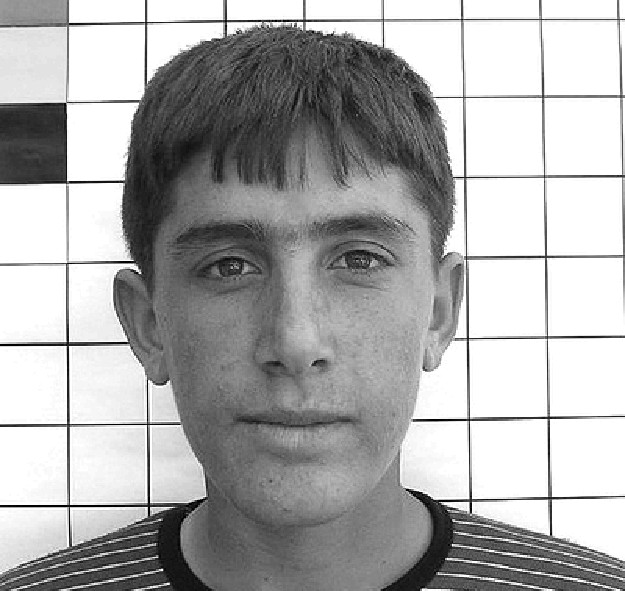} &
			\includegraphics[height = 0.15\textwidth]{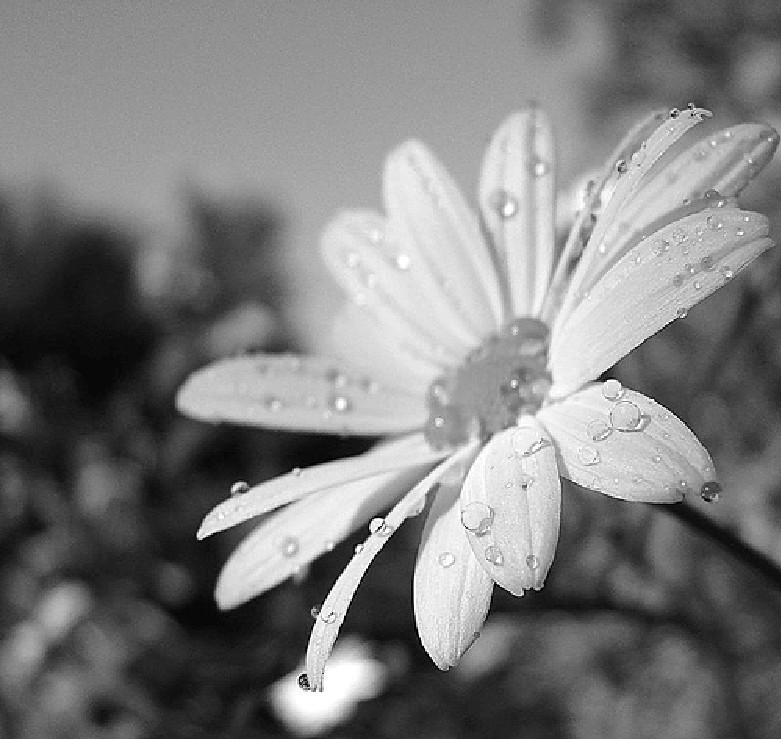} &
			\includegraphics[height = 0.15\textwidth]{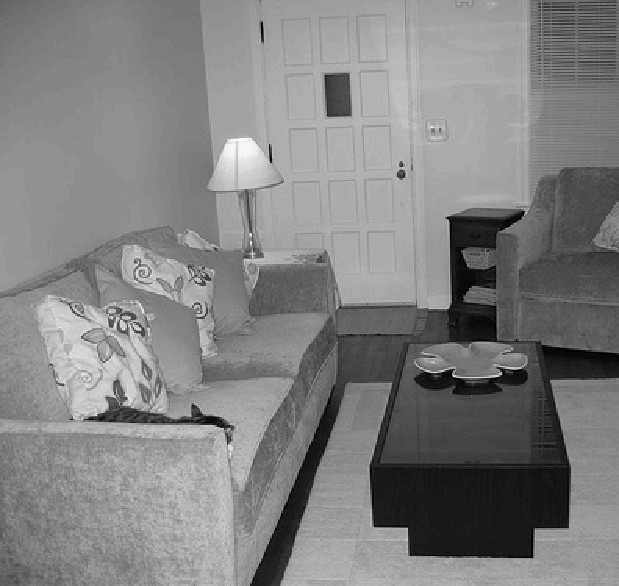} &
			\includegraphics[height = 0.15\textwidth]{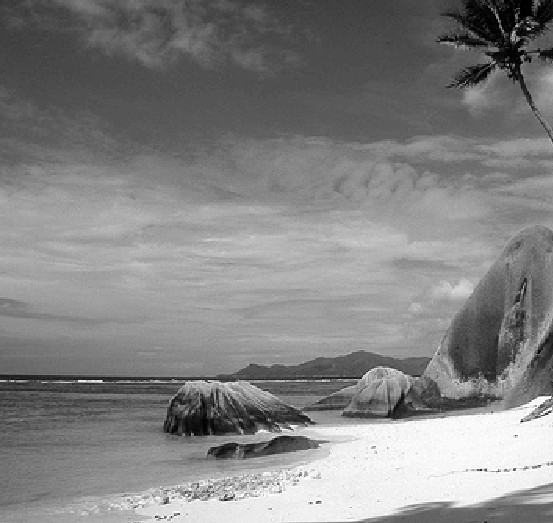} &\\  
			
			\hspace{-1mm}\parbox[b][4em][s]{0.16\textwidth}{Noisy input}&       
			\includegraphics[height = 0.15\textwidth]{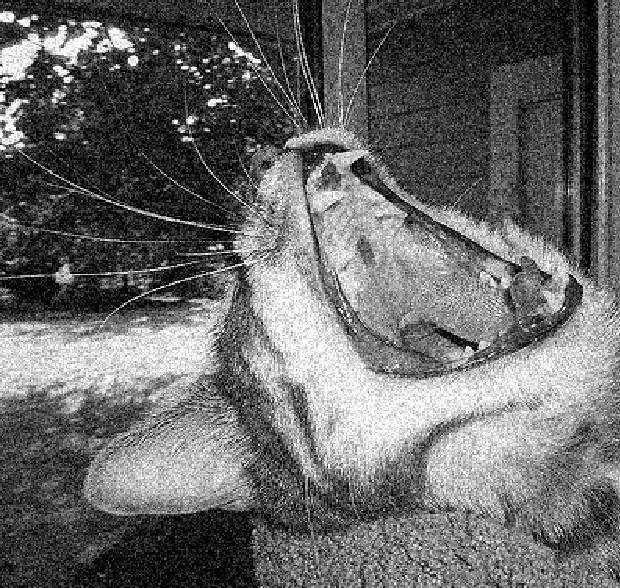} &
			\includegraphics[height = 0.15\textwidth]{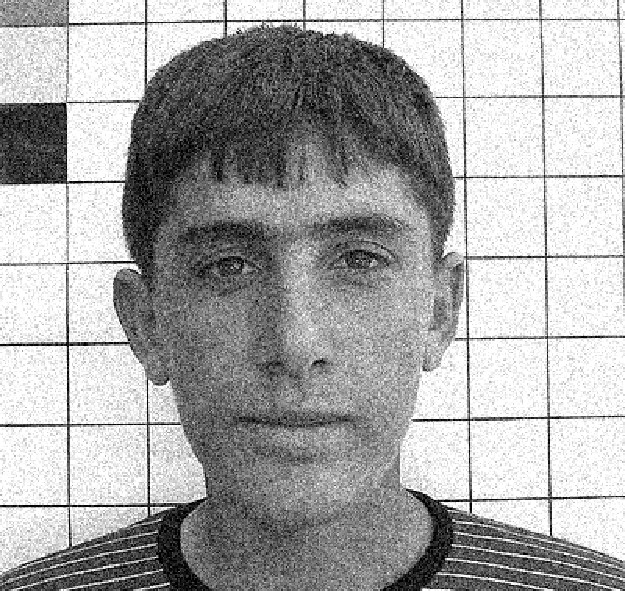} &
			\includegraphics[height = 0.15\textwidth]{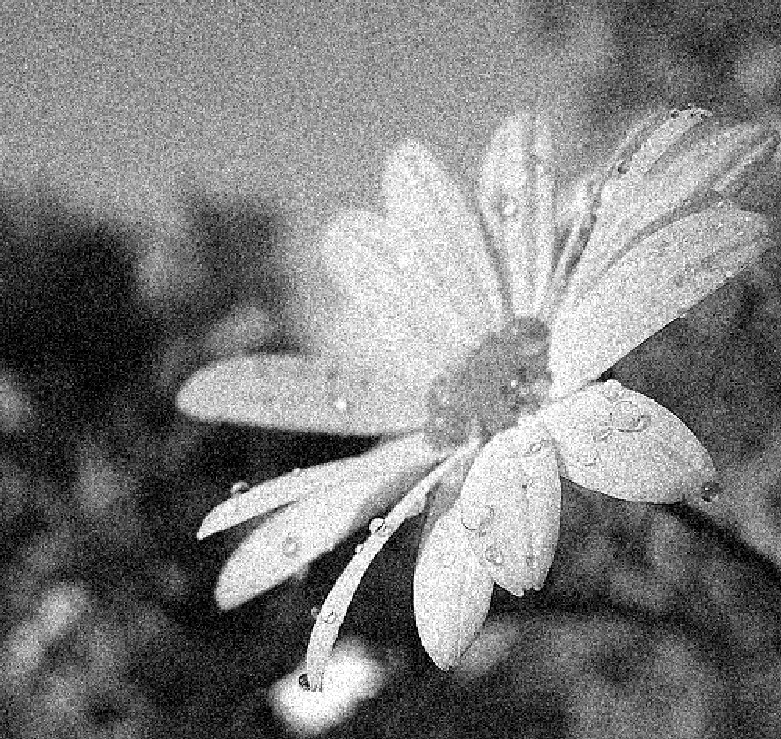} &
			\includegraphics[height = 0.15\textwidth]{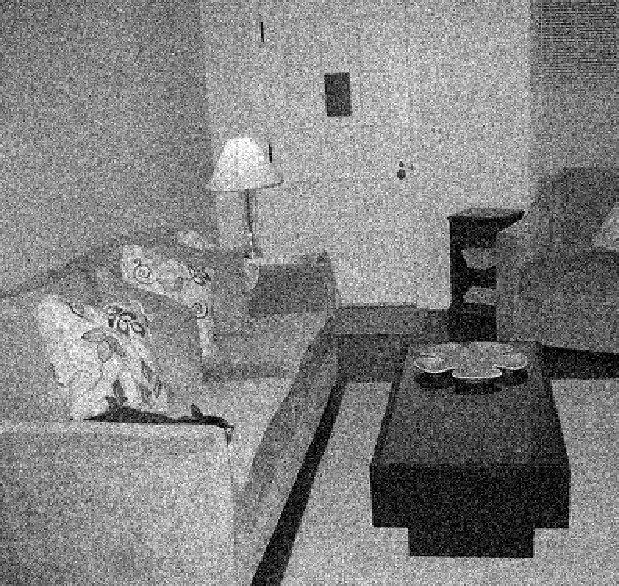} &
			\includegraphics[height = 0.15\textwidth]{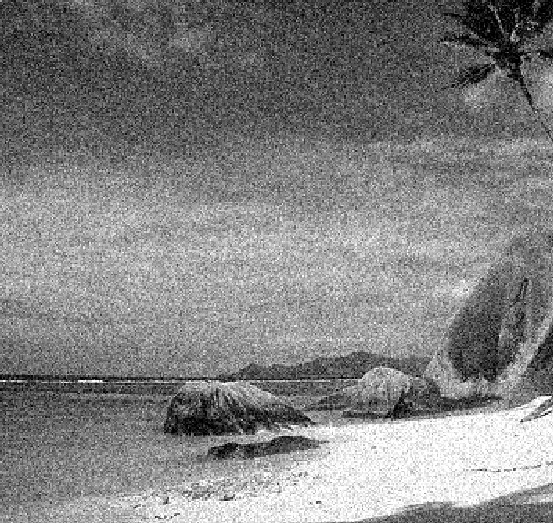} &\\  
			
			\hspace{-1mm}\parbox[b][4em][s]{0.16\textwidth}{Denoised image}&        
			\includegraphics[height = 0.15\textwidth]{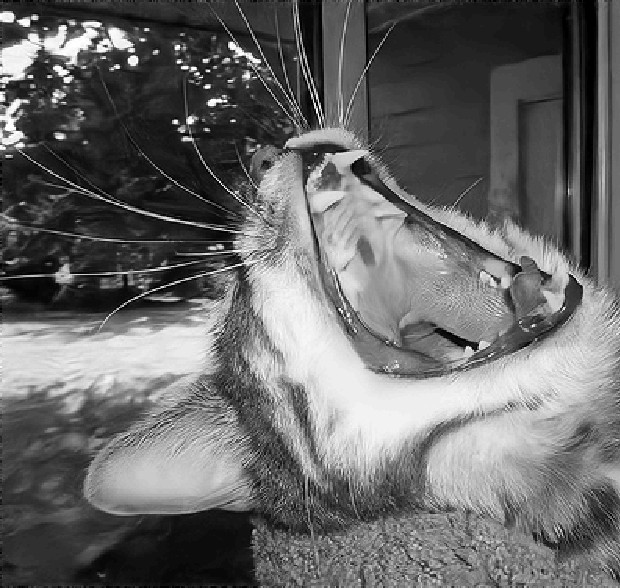} &
			\includegraphics[height = 0.15\textwidth]{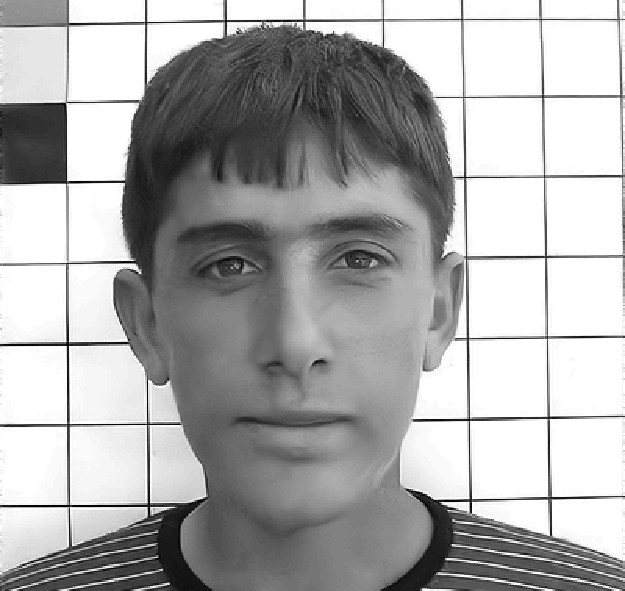} &
			\includegraphics[height = 0.15\textwidth]{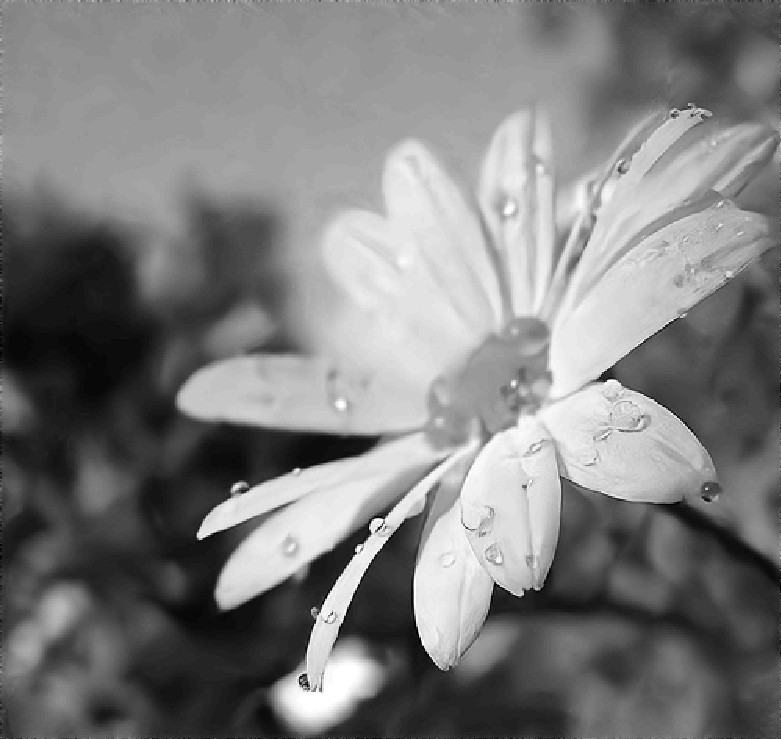} &
			\includegraphics[height = 0.15\textwidth]{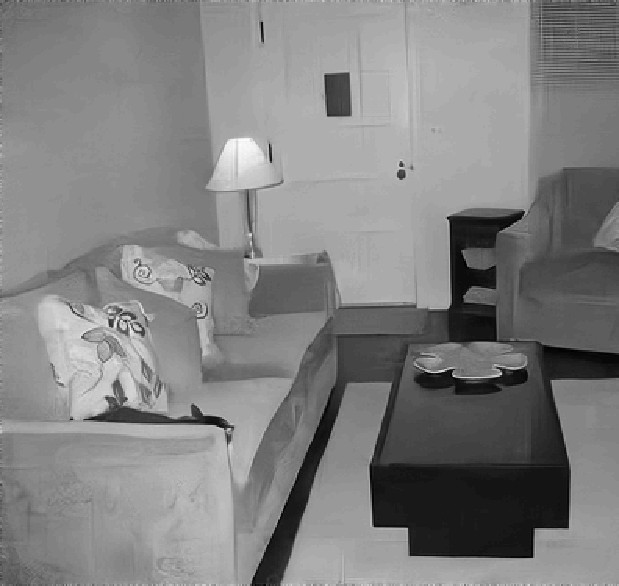} &
			\includegraphics[height = 0.15\textwidth]{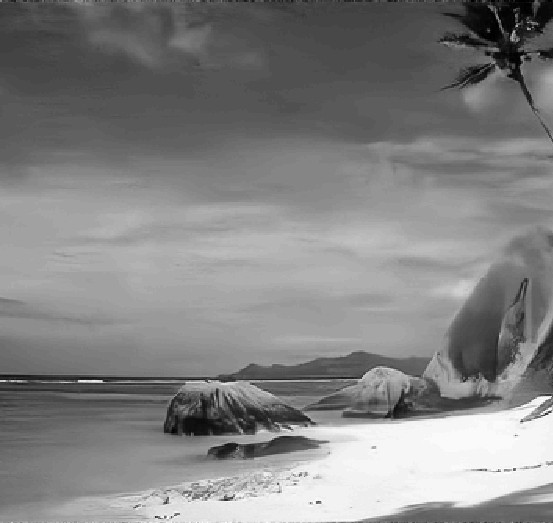} &\\  
			
			\hspace{-1mm}\parbox[b][4em][s]{0.16\textwidth}{Error after 5 layers}&        
			\includegraphics[height = 0.15\textwidth]{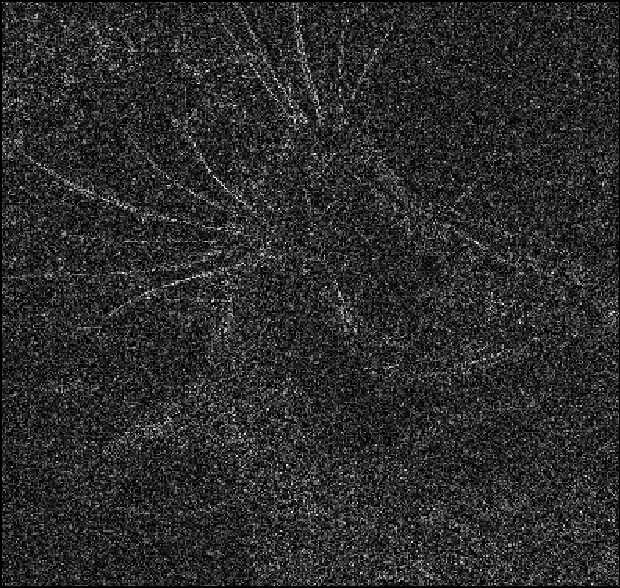} &
			\includegraphics[height = 0.15\textwidth]{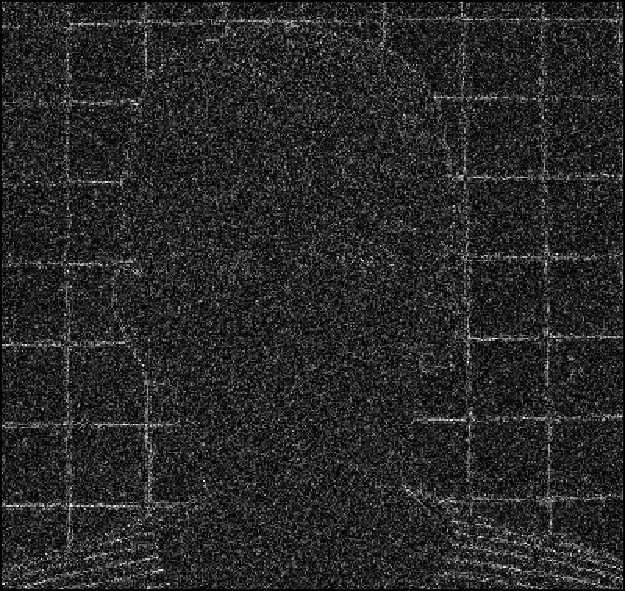} &
			\includegraphics[height = 0.15\textwidth]{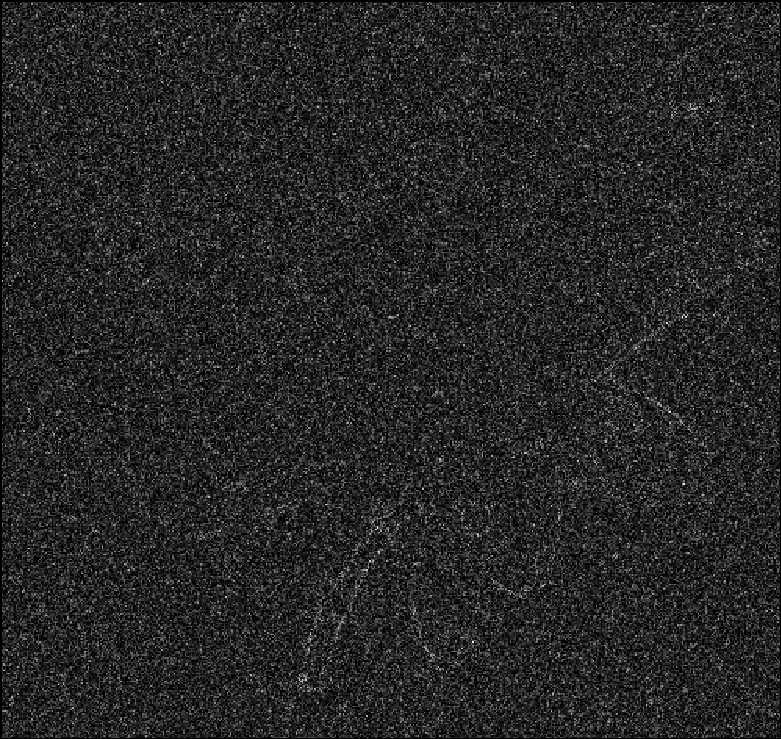} &
			\includegraphics[height = 0.15\textwidth]{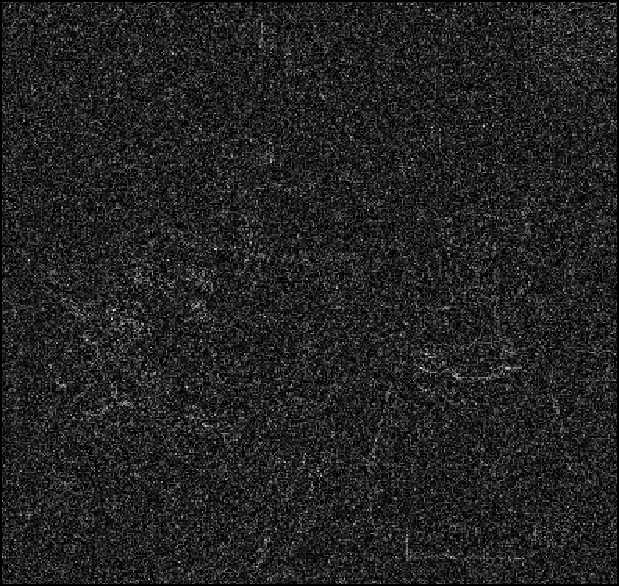} &
			\includegraphics[height = 0.15\textwidth]{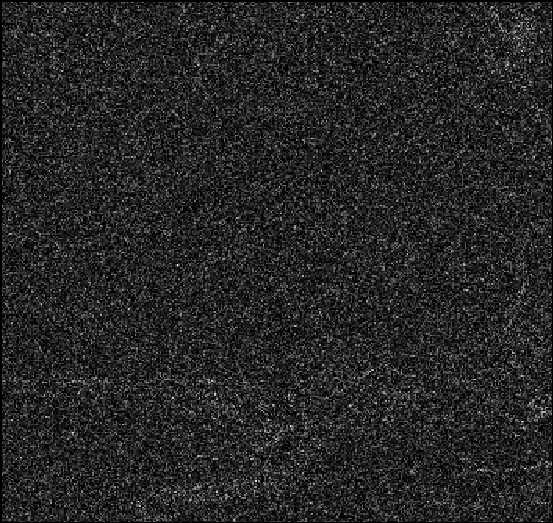} &\\ 
			
			\hspace{-1mm}\parbox[b][4em][s]{0.16\textwidth}{Error after 10 layers}&       
			\includegraphics[height = 0.15\textwidth]{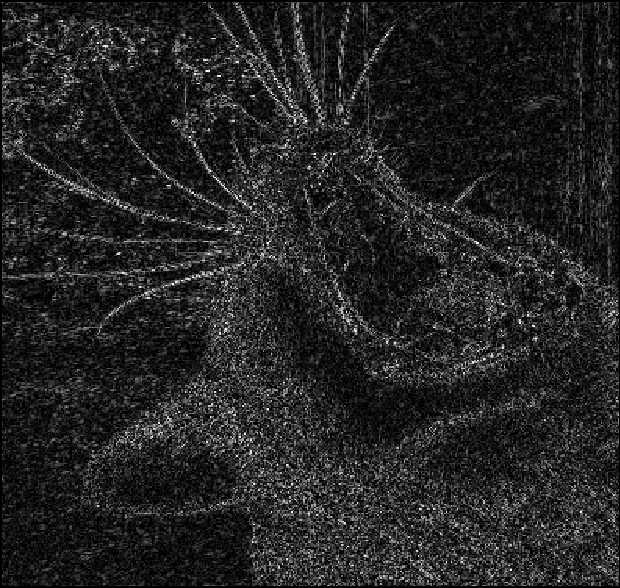} &
			\includegraphics[height = 0.15\textwidth]{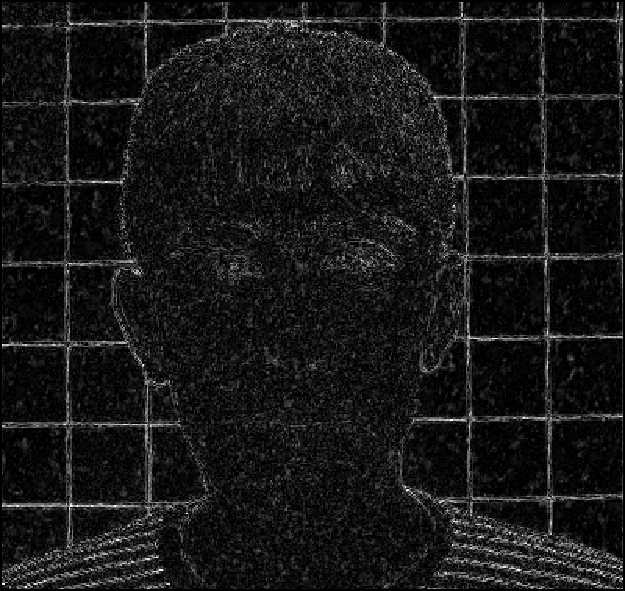} &
			\includegraphics[height = 0.15\textwidth]{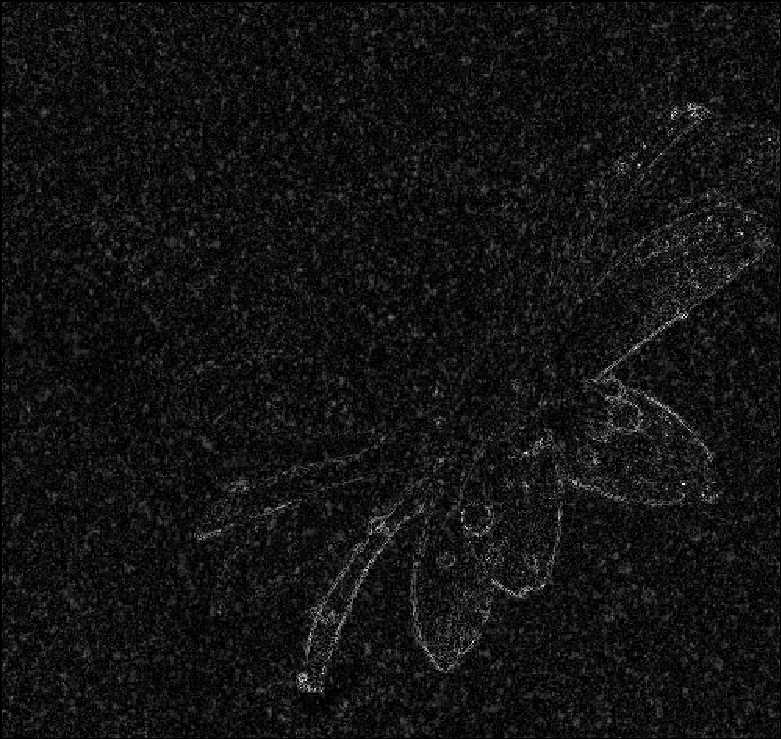} &
			\includegraphics[height = 0.15\textwidth]{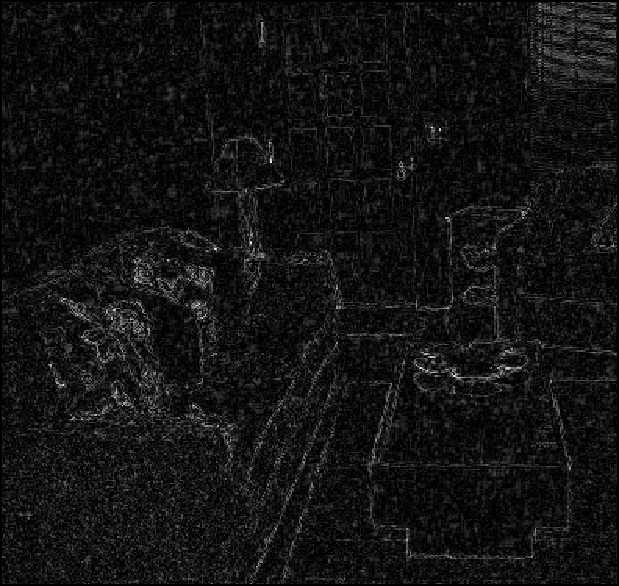} &
			\includegraphics[height = 0.15\textwidth]{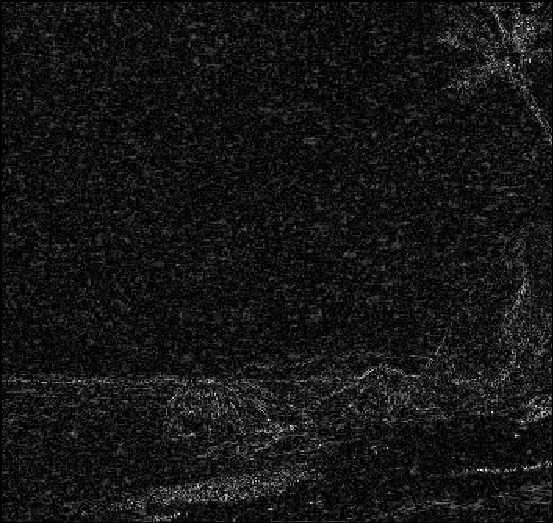} &\\ 
			
			\hspace{-1mm}\parbox[b][4em][s]{0.16\textwidth}{Error after 20 layers (output)}&
			\includegraphics[height = 0.15\textwidth]{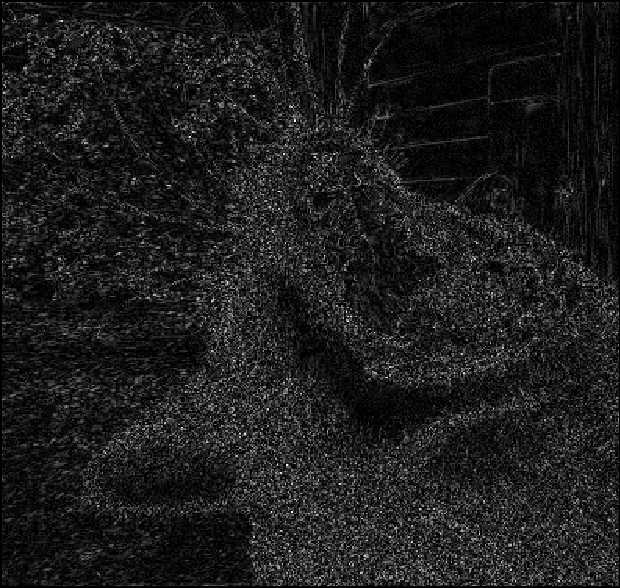} &
			\includegraphics[height = 0.15\textwidth]{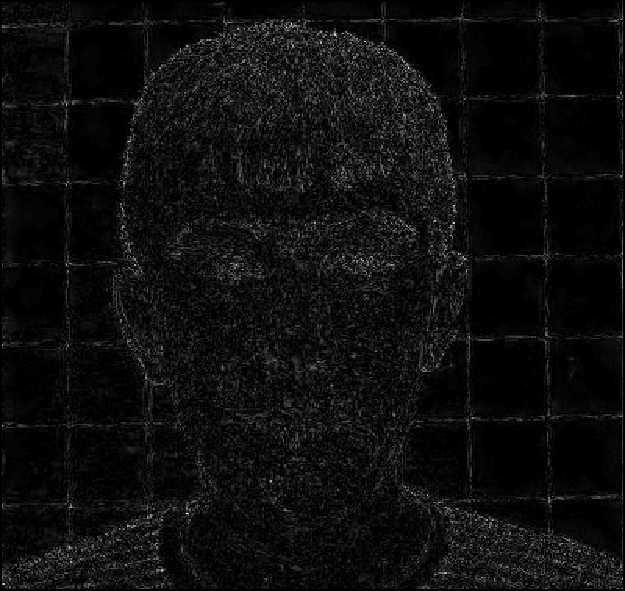} &
			\includegraphics[height = 0.15\textwidth]{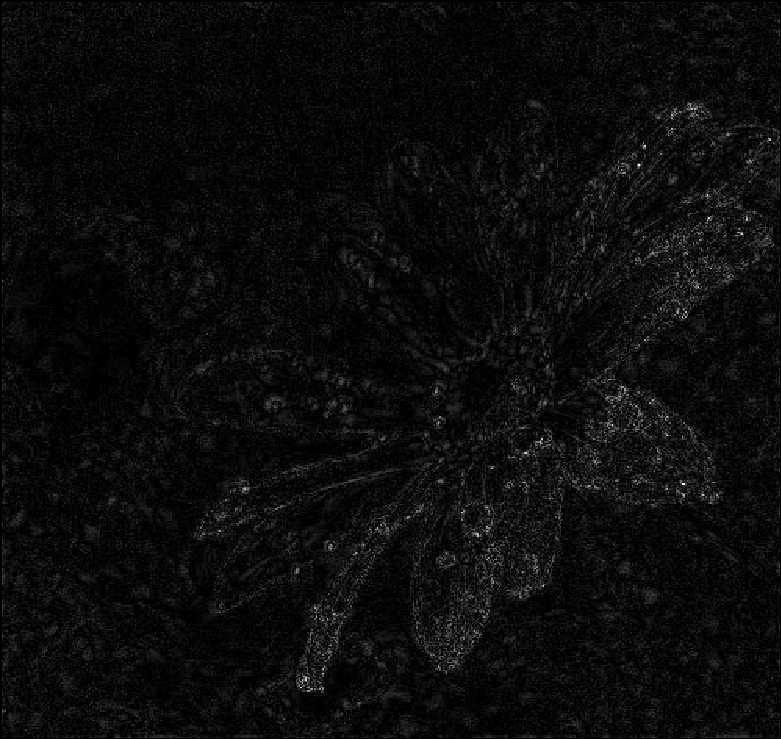} &
			\includegraphics[height = 0.15\textwidth]{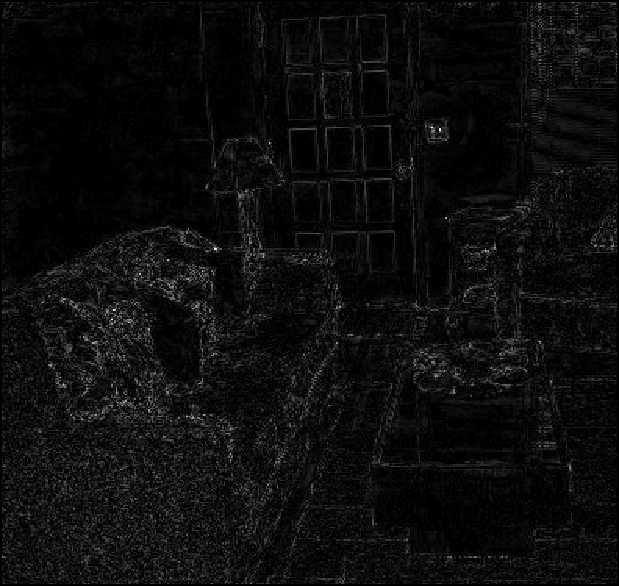} &
			\includegraphics[height = 0.15\textwidth]{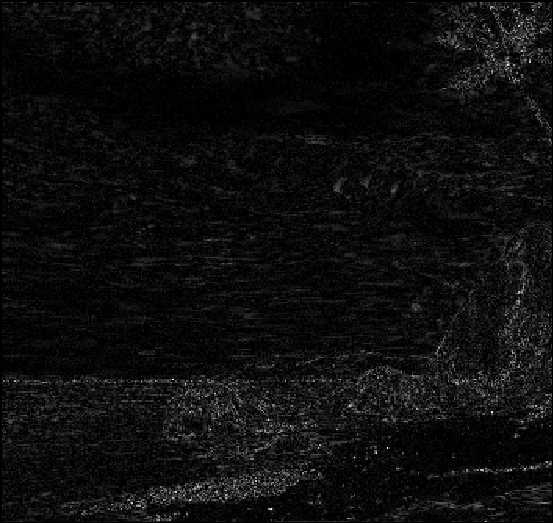} &\\ 
			
			\hspace{-1mm}\parbox[b][4em][s]{0.16\textwidth}{RMSE at different layers}&     
			\includegraphics[width = 0.15\textwidth]{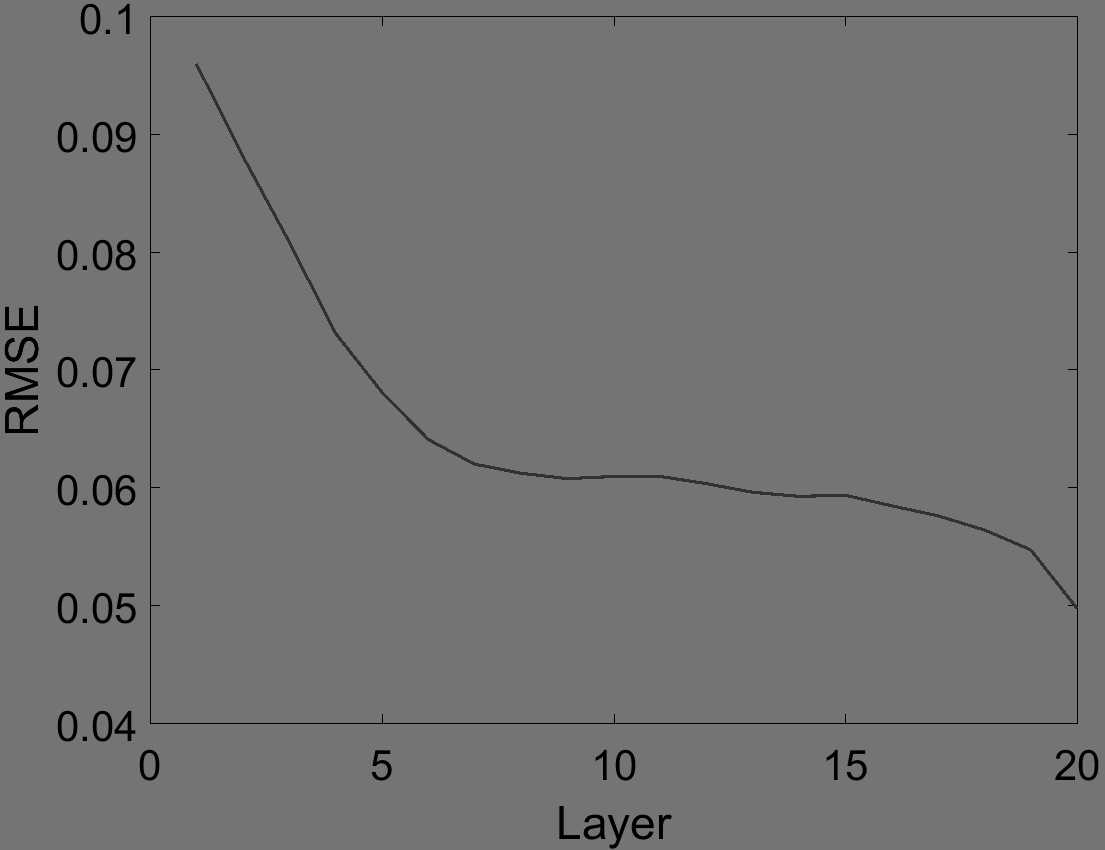}&
			\includegraphics[width = 0.15\textwidth]{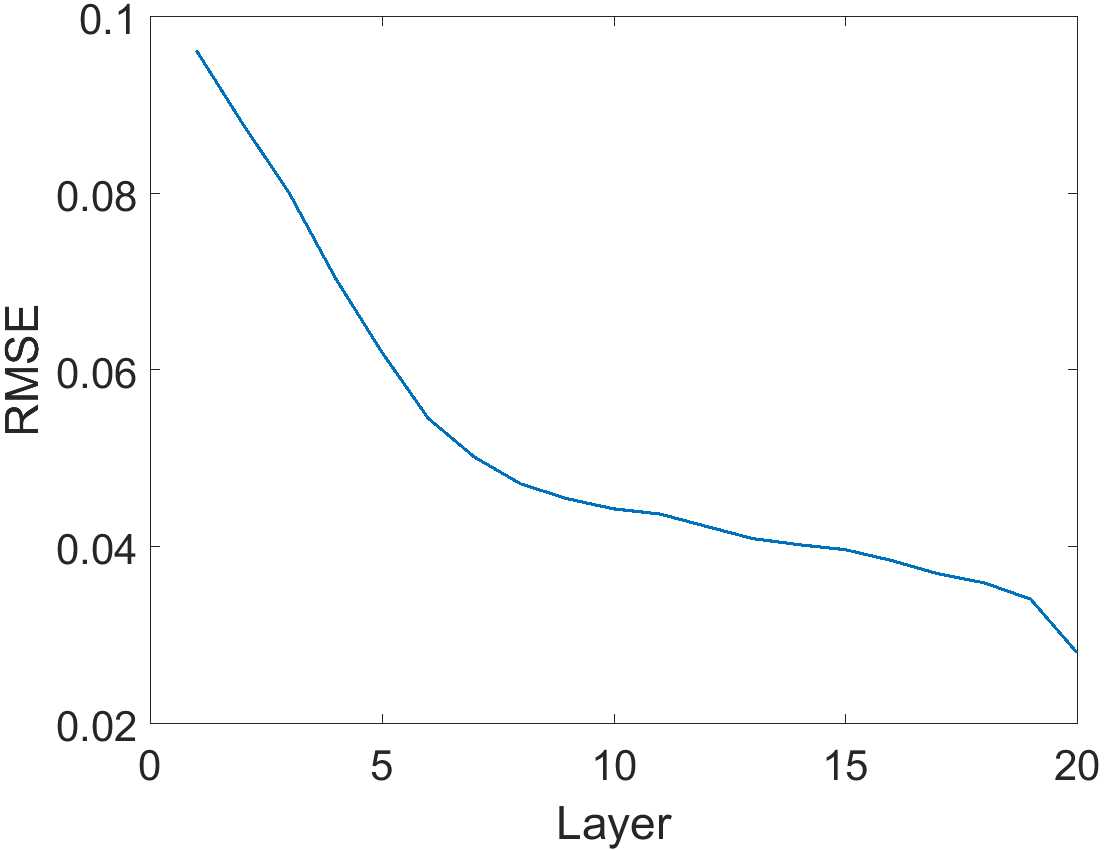}&
			\includegraphics[width = 0.15\textwidth]{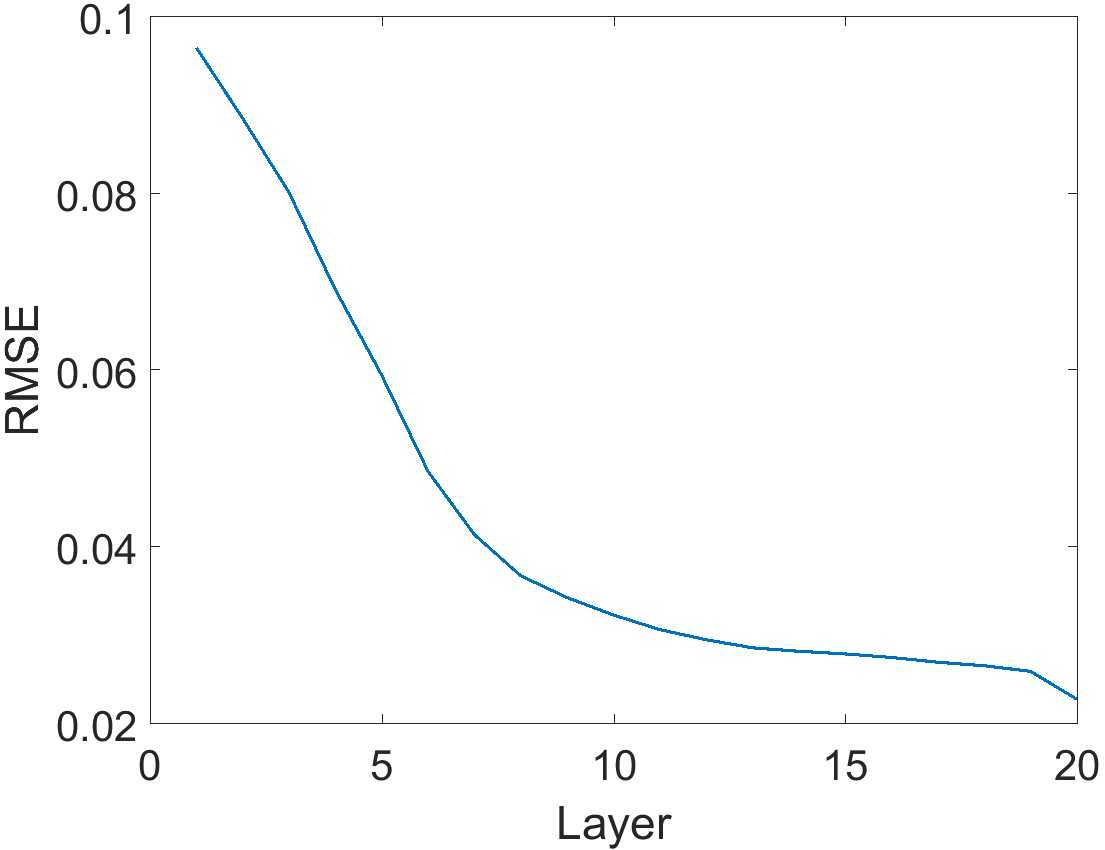}&
			\includegraphics[width = 0.15\textwidth]{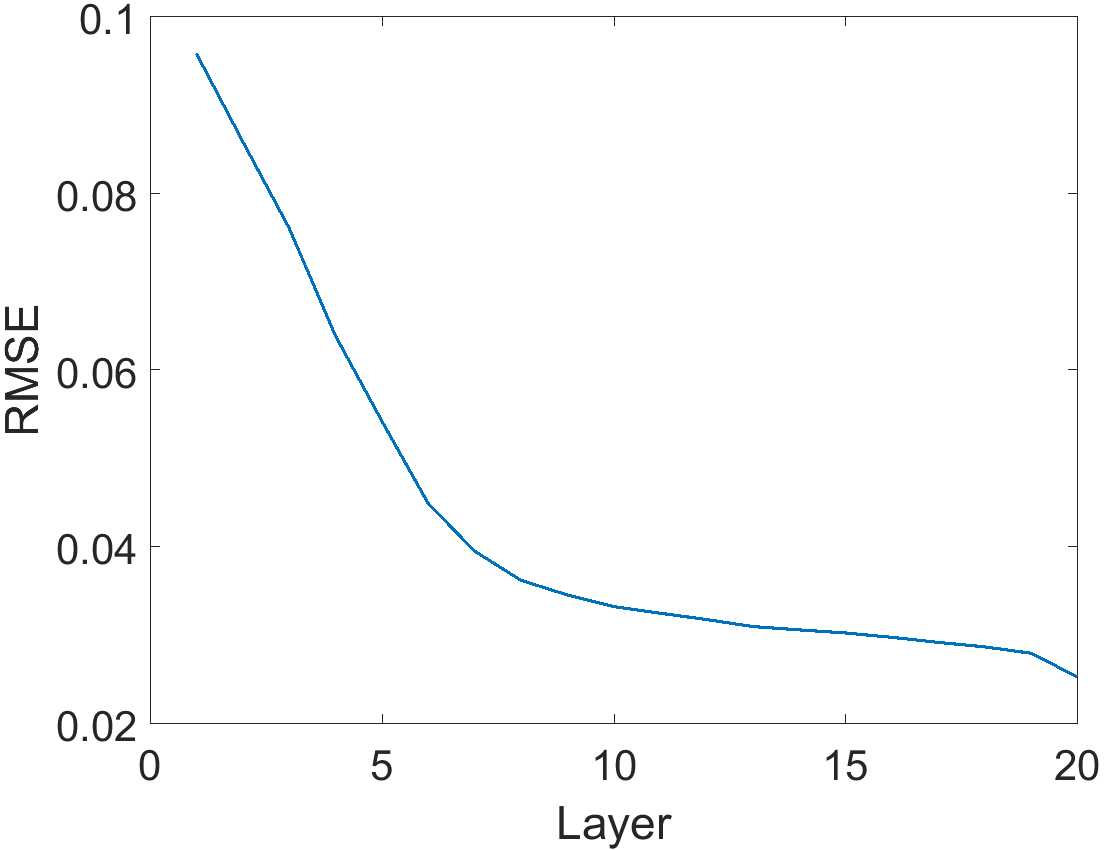}&
			\includegraphics[width = 0.15\textwidth]{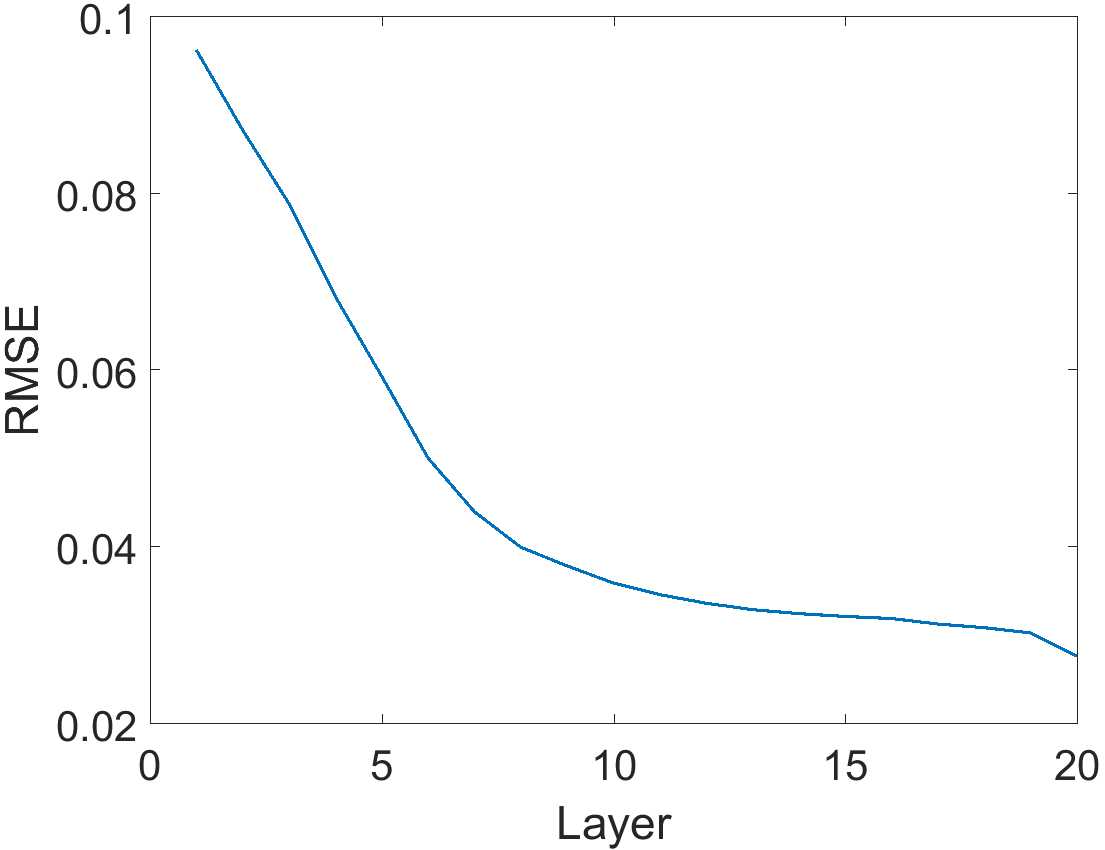}&\\         
			
			\hspace{-1mm}\parbox[b][4em][s]{0.16\textwidth}{Layer contributing the most to each pixel}&
			\includegraphics[height = 0.15\textwidth]{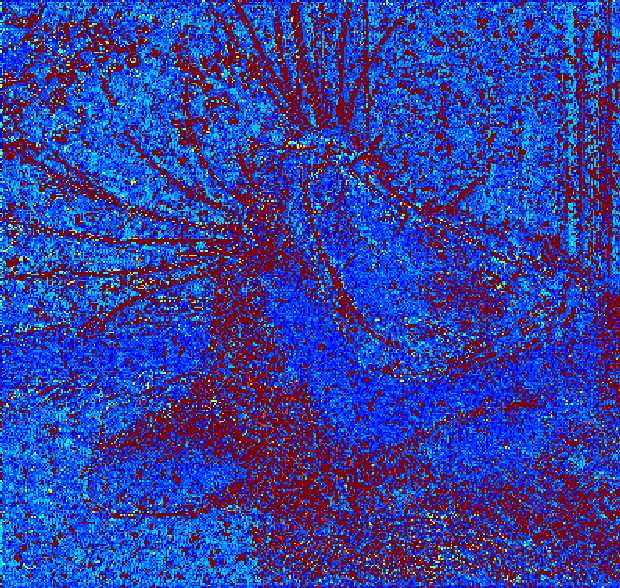} &
			\includegraphics[height = 0.15\textwidth]{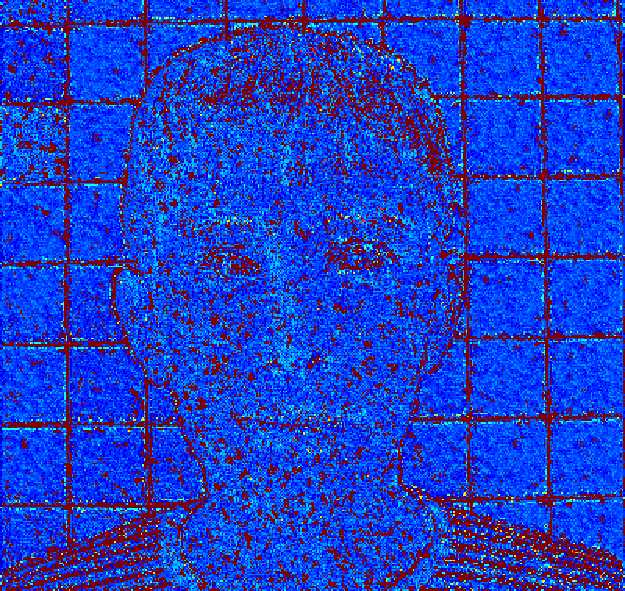} &
			\includegraphics[height = 0.15\textwidth]{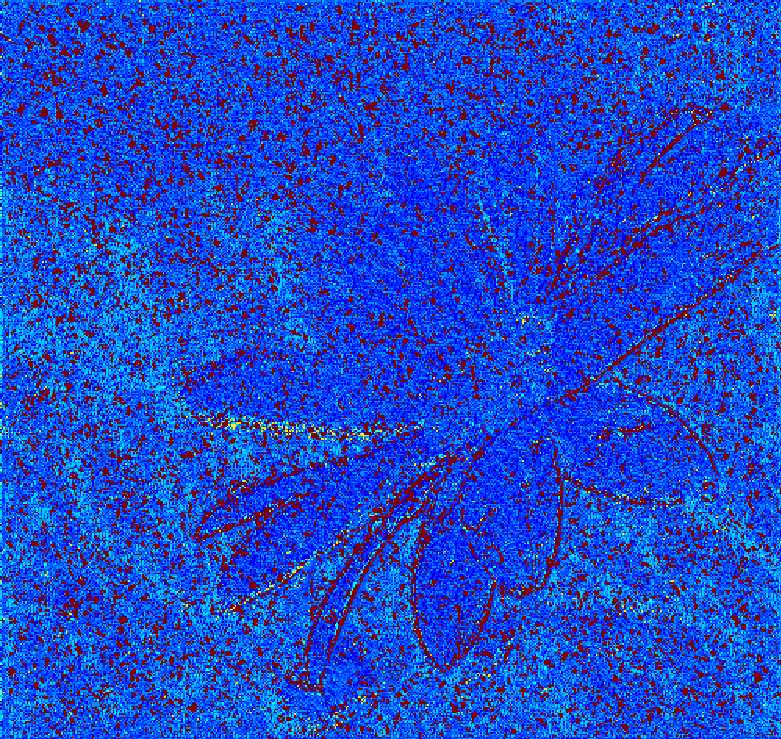} &
			\includegraphics[height = 0.15\textwidth]{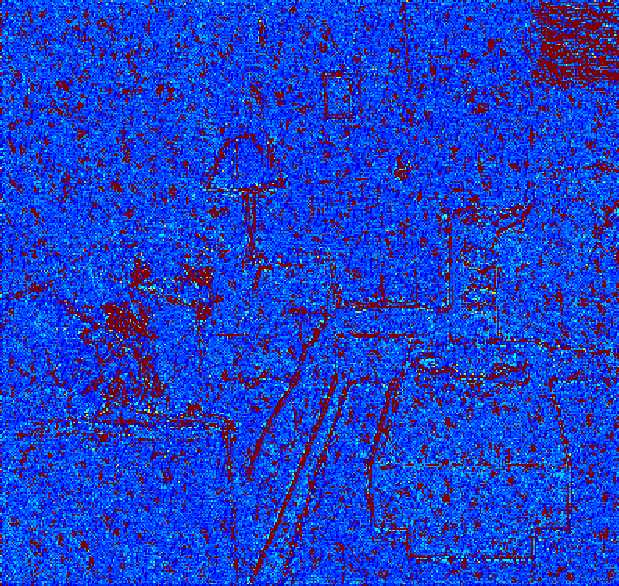} &
			\includegraphics[height = 0.15\textwidth]{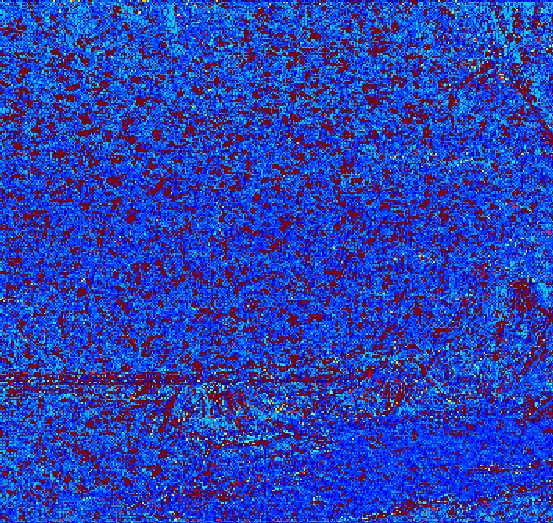} &
			\hspace{-1.5mm} \includegraphics[width = 0.018\textwidth]{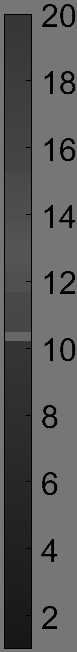} \\

		\end{tabular}   \\
	\end{centering}
	\smallskip
	\caption{\small \textbf{Gradual denoising process. } Images are best viewed electronically, the reader is encouraged to zoom in for a better view. Please refer to Section \ref{sec_noise_estimation} for more details.}
	\label{fig_layer_selection}
\end{figure*}

This section presents a few examples that we believe give insights about the noise estimation of our class-aware networks. The overall noise estimation of the network is the sum of the estimates produced by all individual layers. These are presented in the bottom row of Figure \ref{fig_denoise_flow}. Interestingly, they differ significantly from one another. The shallow layer estimations appear to handle local noise while the deeper ones seem to focus on object contours. In the top row, we present the input image after it has been denoised by all layers up to a specific depth. To further examine what is happening ''under the hood'' of our class-aware denoisers, in Figure \ref{fig_layer_selection} we show the error after $5, 10, $ and $20$ layers (rows $4-6$). Surprisingly, even thought it has not been explicitly enforced at training, the error monotonically decreases with the layer depth (see plots in row $7$). This non-trivial behavior is consistently produced by the network on all test images. Lastly, to visualize which of the layers was the most dominant in the denoising process, we assign a different color to each layer and color each pixel according to the layer in which its value changed the most. The resulting image is shown in the bottom row of Figure \ref{fig_layer_selection}. It can be observed that the first few layers govern the majority of the pixels while the following ones mainly focus on recovering and enhancing the edges and textures that might have been degraded by the first layers.
%


\section{Discussion}
Given the state-of-the-art performance of our network, an important task is to interpret what it has learned and what is the relation between the action of DenoiseNet and the principles governing the previous manually designed state-of-the-art denoising algorithms. 
%
%
%
One such principle that has been shown to improve denoising in recent years is \emph{gradual denoising}, namely that iteratively removing small portions of the noise is preferable to removing it all at once \cite{Romano15Boosting, Sulam15Expected, Zoran11From}. Interestingly, as can be seen in Figure~\ref{fig_denoise_flow}, our network exhibits such a behavior despite the fact it has not been trained explicitly to have a monotonically decreasing error throughout the layers. Each layer in the network removes part of the noise in the image, where the flat regions are being denoised mainly in the first layers, while the edges in the last ones. This may be explained by the fact that the deeper layers corresponds to a larger receptive field and therefore may recover in a better way global patterns such as edges that may be indistinguishable from noise if viewed just in the context of a small patch.

%


In a certain sense, the present research demonstrates that in some cases the whole is \textit{smaller} than the sum of its parts. That is, splitting the input image to several categories and then building a fine-tuned filter for each is preferable over a universal filter. That said, the decision to split according to a semantic class was made due to the immediate availability of off-the-shelf classifiers and their resilience to noise. Yet, this splitting scheme may very well be sub-optimal. Other choices for data partitioning could be made. In particular, a classifier could be learned automatically, e.g., by incorporating the splitting scheme into a network architecture and training it end-to-end. In such cases, the partitioning would lose its simple interpretation as semantic classes, and would instead yield some abstract classes. We defer this interesting direction to future research.

\clearpage
{\small
	\bibliographystyle{ieee}
	\bibliography{egbib}

\begin{thebibliography}{10}\itemsep=-1pt

\bibitem{abadi2015tensorflow}
M.~Abadi, A.~Agarwal, P.~Barham, E.~Brevdo, Z.~Chen, C.~Citro, G.~S. Corrado,
  A.~Davis, J.~Dean, M.~Devin, et~al.
\newblock Tensorflow: Large-scale machine learning on heterogeneous systems,
  2015.
\newblock {\em Software available from tensorflow. org}, 1, 2015.

\bibitem{Aharon06KSVD}
M.~Aharon, M.~Elad, and A.~Bruckstein.
\newblock K-svd: An algorithm for designing overcomplete dictionaries for
  sparse representation.
\newblock {\em IEEE Trans. Signal Process.}, 54(11):4311--4322, Nov 2006.

\bibitem{Anwar15Class}
S.~Anwar, C.~P. Huynh, and F.~Porikli.
\newblock Class-specific image deblurring.
\newblock In {\em IEEE International Conference on Computer Vision (ICCV)},
  pages 495--503, Dec. 2015.

\bibitem{baker2002limits}
S.~Baker and T.~Kanade.
\newblock Limits on super-resolution and how to break them.
\newblock {\em IEEE Transactions on Pattern Analysis and Machine Intelligence},
  24(9):1167--1183, 2002.

\bibitem{Bellegarda16State}
J.~Bellegarda and C.~Monz.
\newblock State of the art in statistical methods for language and speech
  processing.
\newblock {\em Computer Speech and Language}, 35:163–184, Jan. 2016.

\bibitem{Bengio09Learning}
Y.~Bengio.
\newblock Learning deep architectures for ai.
\newblock {\em Foundations and Trends in Machine Learning}, 2(1):1–127, 2009.

\bibitem{Bruna16Super}
J.~Bruna, P.~Sprechmann, and Y.~LeCun.
\newblock Super-resolution with deep convolutional sufficient statistics.
\newblock In {\em ICLR}, 2016.

\bibitem{Bryt08Compression}
O.~Bryt and M.~Elad.
\newblock Compression of facial images using the {K-SVD} algorithm.
\newblock {\em Journal of Visual Communication and Image Representation},
  19(4):270 -- 282, 2008.

\bibitem{Buades05Non}
A.~Buades, B.~Coll, , and J.~Morel.
\newblock A non-local algorithm for image denoising.
\newblock In {\em IEEE Conference on Computer Vision and Pattern Recognition
  (CVPR)}, 2005.

\bibitem{burger2012image}
H.~C. Burger, C.~J. Schuler, and S.~Harmeling.
\newblock Image denoising: Can plain neural networks compete with bm3d?
\newblock In {\em IEEE Conference on Computer Vision and Pattern Recognition
  (CVPR)}, pages 2392--2399. IEEE, 2012.

\bibitem{Chan16Plug}
S.~Chan, X.~Wang, and O.~Elgendy.
\newblock Plug-and-play admm for image restoration: Fixed point convergence and
  applications.
\newblock {\em ArXiv, abs/1605.01710}, 2016.

\bibitem{Chatterjee10Denoising}
P.~Chatterjee and P.~Milanfar.
\newblock Is denoising dead?
\newblock {\em IEEE Trans. Image Process.}, 19(4):895–911, 2010.

\bibitem{Chen16Trainable}
Y.~Chen and T.~Pock.
\newblock Trainable nonlinear reaction diffusion: A flexible framework for fast
  and effective image restoration.
\newblock {\em IEEE Transactions on Pattern Analysis and Machine Intelligence
  (CVPR)}, 2016.

\bibitem{dabov2007image}
K.~Dabov, A.~Foi, V.~Katkovnik, and K.~Egiazarian.
\newblock Image denoising by sparse 3-d transform-domain collaborative
  filtering.
\newblock {\em IEEE Trans. Image Process.}, 16(8):2080--2095, 2007.

\bibitem{Dar16Postprocessing}
Y.~Dar, A.~M. Bruckstein, M.~Elad, and R.~Giryes.
\newblock Postprocessing of compressed images via sequential denoising.
\newblock {\em IEEE Trans. Imag. Proc.}, 25(7):3044--3058, 2016.

\bibitem{Delbracio15Burst}
M.~Delbracio and G.~Sapiro.
\newblock Burst deblurring: Removing camera shake through fourier burst
  accumulation.
\newblock In {\em IEEE Conference on Computer Vision and Pattern Recognition
  (CVPR)}, 2015.

\bibitem{Deng14Deep}
L.~Deng and D.~Yu.
\newblock Deep learning: Methods and applications.
\newblock {\em Foundations and Trends in Signal Processing}, 7(3-4):197–387,
  2014.

\bibitem{Dong15Image}
W.~Dong, G.~Shi, Y.~Ma, and X.~Li.
\newblock Image restoration via simultaneous sparse coding: Where structured
  sparsity meets gaussian scale mixture.
\newblock {\em International Journal of Computer Vision (IJCV)},
  114(2):217--232, Sep.

\bibitem{Dong13Nonlocally}
W.~Dong, L.~Zhang, G.~Shi, and X.~Li.
\newblock Nonlocally centralized sparse representation for image restoration.
\newblock {\em IEEE Trans. Image Process.}, 22(4):1620--1630, April 2013.

\bibitem{everingham2010pascal}
M.~Everingham, L.~Van~Gool, C.~K. Williams, J.~Winn, and A.~Zisserman.
\newblock The pascal visual object classes (voc) challenge.
\newblock {\em International journal of computer vision}, 88(2):303--338, 2010.

\bibitem{pascal-voc-2010}
M.~Everingham, L.~Van~Gool, C.~K.~I. Williams, J.~Winn, and A.~Zisserman.
\newblock The {PASCAL} {V}isual {O}bject {C}lasses {C}hallenge 2010 {(VOC2010)}
  {R}esults.
\newblock
  http://www.pascal-network.org/challenges/VOC/voc2010/workshop/index.html.

\bibitem{pascal-voc-2012}
M.~Everingham, L.~Van~Gool, C.~K.~I. Williams, J.~Winn, and A.~Zisserman.
\newblock The {PASCAL} {V}isual {O}bject {C}lasses {C}hallenge 2012 {(VOC2012)}
  {R}esults.
\newblock
  http://www.pascal-network.org/challenges/VOC/voc2012/workshop/index.html.

\bibitem{Goodfellow16Deep}
I.~Goodfellow, Y.~Bengio, and A.~Courville.
\newblock {\em Deep learning}.
\newblock Book in preparation for MIT Press, 2016.

\bibitem{Greenspan16Guest}
H.~Greenspan, B.~van Ginneken, and R.~M. Summers.
\newblock Guest editorial deep learning in medical imaging: Overview and future
  promise of an exciting new technique.
\newblock {\em IEEE Transactions on Medical Imaging}, 35(5):1153--1159, May
  2016.

\bibitem{He16Deep}
K.~He, X.~Zhang, S.~Ren, and J.~Sun.
\newblock Deep residual learning for image recognition.
\newblock In {\em IEEE Conference on Computer Vision and Pattern Recognition
  (CVPR)}, 2016.

\bibitem{Hirschberg15Advances}
J.~Hirschberg and C.~D. Manning.
\newblock Advances in natural language processing.
\newblock {\em Science}, 349(6245):261--266, 2015.

\bibitem{Iizuka16Let}
S.~Iizuka, E.~Simo-Serra, and H.~Ishikawa.
\newblock Let there be color!: Joint end-to-end learning of global and local
  image priors for automatic image colorization with simultaneous
  classification.
\newblock In {\em SIGGRAPH}, 2016.

\bibitem{Joshi10Personal}
N.~Joshi, W.~Matusik, E.~H. Adelson, and D.~J. Kriegman.
\newblock Personal photo enhancement using example images.
\newblock {\em ACM Trans. Graph.}, 29(2):12:1--12:15, Apr. 2010.

\bibitem{Karpathy14Large}
A.~Karpathy, G.~Toderici, S.~Shetty, T.~Leung, R.~Sukthankar, and L.~Fei-Fei.
\newblock Large-scale video classification with convolutional neural networks.
\newblock In {\em CVPR}, 2014.

\bibitem{Kim16Accurate}
J.~Kim, J.~K. Lee, and K.~M. Lee.
\newblock Accurate image super-resolution using very deep convolutional
  networks.
\newblock In {\em CVPR}, 2016.

\bibitem{DBLP:journals/corr/KingmaB14}
D.~P. Kingma and J.~Ba.
\newblock Adam: {A} method for stochastic optimization.
\newblock In {\em ICLR}, 2015.

\bibitem{Krizhevsky12ImageNet}
A.~Krizhevsky, I.~Sutskever, and G.~E. Hinton.
\newblock Imagenet classification with deep convolutional neural networks.
\newblock In {\em Advances in Neural Information Processing Systems (NIPS)},
  pages 1097--1105, 2012.

\bibitem{Lebrun13Nonlocal}
M.~Lebrun, A.~Buades, and J.~M. Morel.
\newblock A nonlocal bayesian image denoising algorithm.
\newblock {\em SIAM Journal on Imaging Sciences}, 6(3):1665--1688, 2013.

\bibitem{LeCun15Deep}
Y.~LeCun, Y.~Bengio, and G.~Hinton.
\newblock Deep learning.
\newblock {\em Foundations and Trends in Signal Processing}, 521:436–444, May
  2015.

\bibitem{Ledig16Photo}
C.~Ledig, L.~Theis, F.~Husz{\'a}r, J.~Caballero, A.~Aitken, A.~Tejani, J.~Totz,
  Z.~Wang, and W.~Shi.
\newblock Photo-realistic single image super-resolution using a generative
  adversarial network.
\newblock {\em arXiv abs/1609.04802}, 2016.

\bibitem{levin2011natural}
A.~Levin and B.~Nadler.
\newblock Natural image denoising: Optimality and inherent bounds.
\newblock In {\em IEEE Conference on Computer Vision and Pattern Recognition
  (CVPR)}, pages 2833--2840. IEEE, 2011.

\bibitem{Levin12Patch}
A.~Levin, B.~Nadler, F.~Durand, and W.~Freeman.
\newblock Patch complexity, finite pixel correlations and optimal denoising.
\newblock In {\em ECCV}, 2012.

\bibitem{Mairal09Non}
J.~Mairal, F.~Bach, J.~Ponce, G.~Sapiro, and A.~Zisserman.
\newblock Non-local sparse models for image restoration.
\newblock In {\em ICCV}, pages 2272--2279, 2009.

\bibitem{Makitalo11Optimal}
M.~Makitalo and A.~Foi.
\newblock Optimal inversion of the {Anscombe} transformation in low-count
  {Poisson} image denoising.
\newblock {\em IEEE Trans. on Image Proces.}, 20(1):99--109, Jan. 2011.

\bibitem{Makitalo14Noise}
M.~Makitalo and A.~Foi.
\newblock Noise parameter mismatch in variance stabilization, with an
  application to poisson-gaussian noise estimation.
\newblock {\em IEEE Trans. on Image Proces.}, 23(12):5348--5359, Jan. 2014.

\bibitem{MartinFTM01}
D.~Martin, C.~Fowlkes, D.~Tal, and J.~Malik.
\newblock A database of human segmented natural images and its application to
  evaluating segmentation algorithms and measuring ecological statistics.
\newblock In {\em Proc. 8th Int'l Conf. Computer Vision}, volume~2, pages
  416--423, July 2001.

\bibitem{Mnih15Human}
V.~Mnih, K.~Kavukcuoglu, D.~Silver, A.~A. Rusu, J.~Veness, M.~G. Bellemare,
  A.~Graves, M.~Riedmiller, A.~K. Fidjeland, G.~Ostrovski, S.~Petersen,
  C.~Beattie, A.~Sadik, I.~Antonoglou, H.~King, D.~Kumaran, D.~Wierstra,
  S.~Legg, and D.~Hassabis.
\newblock Human-level control through deep reinforcement learning.
\newblock {\em Nature}, 518:529–533, Feb. 2015.

\bibitem{Pathak16Context}
D.~Pathak, P.~Krahenbuhl, J.~Donahue, T.~Darrell, and A.~A. Efros.
\newblock Context encoders: Feature learning by inpainting.
\newblock In {\em CVPR}, 2016.

\bibitem{Poznanski16CNN}
A.~Poznanski and L.~Wolf.
\newblock Cnn-n-gram for handwriting word recognition.
\newblock In {\em CVPR}, 2016.

\bibitem{Romano15Boosting}
Y.~Romano and M.~Elad.
\newblock Boosting of image denoising algorithms.
\newblock {\em SIAM Journal on Imaging Sciences}, 8(2):1187--1219, 2015.

\bibitem{Romano16Little}
Y.~Romano, M.~Elad, and P.~Milanfar.
\newblock The little engine that could: Regularization by denoising (red).
\newblock {\em arXiv:1611.02862}, 2016.

\bibitem{Rond16Poisson}
A.~Rond, R.~Giryes, and M.~Elad.
\newblock Poisson inverse problems by the plug-and-play scheme.
\newblock {\em Journal of Visual Communication and Image Representation}, 2016.

\bibitem{roth2009fields}
S.~Roth and M.~J. Black.
\newblock Fields of experts.
\newblock {\em International Journal of Computer Vision}, 82(2):205--229, 2009.

\bibitem{ImageNet15}
O.~Russakovsky, J.~Deng, H.~Su, J.~Krause, S.~Satheesh, S.~Ma, Z.~Huang,
  A.~Karpathy, A.~Khosla, M.~Bernstein, A.~C. Berg, and L.~Fei-Fei.
\newblock {ImageNet Large Scale Visual Recognition Challenge}.
\newblock {\em Int. Journal of Computer Vision}, 115(3):211--252, 2015.

\bibitem{Sak15Fast}
H.~Sak, A.~Senior, K.~Rao, and F.~Beaufays.
\newblock Fast and accurate recurrent neural network acoustic models for speech
  recognition.
\newblock In {\em INTERSPEECH}, 2015.

\bibitem{Schmidhuber15Deep}
J.~Schmidhuber.
\newblock Deep learning in neural networks: An overview.
\newblock {\em Neural Networks}, 61:85--117, 2015.

\bibitem{Schmidt10Generative}
U.~Schmidt, Q.~Gao, and S.~Roth.
\newblock A generative perspective on mrfs in low-level vision.
\newblock In {\em IEEE Conference on Computer Vision and Pattern Recognition
  (CVPR)}, 2010.

\bibitem{Schroff15FaceNet}
F.~Schroff, D.~Kalenichenko, and J.~Philbin.
\newblock Facenet: A unified embedding for face recognition and clustering.
\newblock In {\em CVPR}, 2015.

\bibitem{Schuler13machine}
C.~J. Schuler, H.~C. Burger, S.~Harmeling, and B.~Schölkopf.
\newblock A machine learning approach for non-blind image deconvolution.
\newblock In {\em CVPR}, 2013.

\bibitem{Silver16Mastering}
D.~Silver, A.~Huang, C.~Maddison, A.~Guez, L.~Sifre, G.~van~den Driessche,
  J.~Schrittwieser, I.~Antonoglou, V.~Panneershelvam, M.~Lanctot, S.~Dieleman,
  D.~Grewe, J.~Nham, N.~Kalchbrenner, I.~Sutskever, T.~Lillicrap, M.~Leach,
  K.~Kavukcuoglu, T.~Graepel, and D.~Hassabis.
\newblock Mastering the game of go with deep neural networks and tree search.
\newblock {\em Nature}, 529:484–489, 2016.

\bibitem{Socher13Recursive}
R.~Socher, A.~Perelygin, J.~Wu, J.~Chuang, C.~Manning, A.~Ng, and C.~Potts.
\newblock Recursive deep models for semantic compositionality over a sentiment
  treebank.
\newblock In {\em EMNLP}, 2013.

\bibitem{Sreehari16Plug}
S.~Sreehari, S.~V. Venkatakrishnan, B.~Wohlberg, G.~T. Buzzard, L.~F. Drummy,
  J.~P. Simmons, and C.~A. Bouman.
\newblock Plug-and-play priors for bright field electron tomography and sparse
  interpolation.
\newblock {\em IEEE Transactions on Computational Imaging}, 2(4):408--423, Dec
  2016.

\bibitem{Sulam15Expected}
J.~Sulam and M.~Elad.
\newblock Expected patch log likelihood with a sparse prior.
\newblock In {\em Energy Minimization Methods in Computer Vision and Pattern
  Recognition (EMMCVPR)}, Hong-Kong, 2015.

\bibitem{Sutskever14Sequence}
I.~Sutskever, O.~Vinyals, and Q.~Le.
\newblock Sequence to sequence learning with neural networks.
\newblock In {\em NIPS}, 2014.

\bibitem{Szegedy15Going}
C.~Szegedy, W.~Liu, Y.~Jia, P.~Sermanet, S.~Reed, D.~Anguelov, D.~Erhan,
  V.~Vanhoucke, and A.~Rabinovich.
\newblock Going deeper with convolutions.
\newblock In {\em CVPR}, 2015.

\bibitem{DBLP:journals/corr/SzegedyVISW15}
C.~Szegedy, V.~Vanhoucke, S.~Ioffe, J.~Shlens, and Z.~Wojna.
\newblock Rethinking the inception architecture for computer vision, journal =
  {arXiv, abs/1512.00567}, year = {2015}, url =
  {http://arxiv.org/abs/1512.00567},.

\bibitem{vemulapalli2016deep}
R.~Vemulapalli, O.~Tuzel, and M.-Y. Liu.
\newblock Deep gaussian conditional random field network: A model-based deep
  network for discriminative denoising.
\newblock In {\em IEEE Conference on Computer Vision and Pattern Recognition
  (CVPR)}, 2016.

\bibitem{Venkatakrishnan13Plug}
S.~Venkatakrishnan, C.~Bouman, and B.~Wohlberg.
\newblock Plug-and-play priors for model based reconstruction.
\newblock In {\em GlobalSIP}, 2013.

\bibitem{Wang14Comprehensive}
N.~Wang, D.~Tao, X.~Gao, X.~Li, and J.~Li.
\newblock A comprehensive survey to face hallucination.
\newblock {\em International Journal of Computer Vision}, 106(1):9--30, 2014.

\bibitem{Yu12Solving}
G.~Yu, G.~Sapiro, and S.~Mallat.
\newblock Solving inverse problems with piecewise linear estimators: From
  {G}aussian mixture models to structured sparsity.
\newblock {\em IEEE Trans. Image Process.}, 21(5):2481 --2499, may 2012.

\bibitem{zhang2016colorful}
R.~Zhang, P.~Isola, and A.~A. Efros.
\newblock Colorful image colorization.
\newblock {\em ECCV}, 2016.

\bibitem{Zoran11From}
D.~Zoran and Y.~Weiss.
\newblock From learning models of natural image patches to whole image
  restoration.
\newblock In {\em ICCV}, 2011.

\end{thebibliography}
}
\clearpage

\section*{Denoising Examples}

In the next few pages we include two examples for each of the six semantic classes used in the paper. For each example, our class-aware denoiser is compared to BM3D [14], MLP [10] and TNRD [13]. In addition we present the ground truth and noisy images. Although in most cases the difference between the methods is visible in full view, we encourage the reader to zoom-in to fully appreciate fine image details.

\begin{figure*}[]
	\centering    
	\begin{tabular}{c@{\hskip 0.005\textwidth}c@{\hskip 0.005\textwidth}c}
		
		\includegraphics[width = 0.33\textwidth]{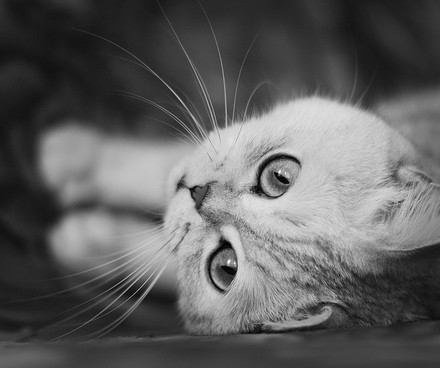} &
		\includegraphics[width = 0.33\textwidth]{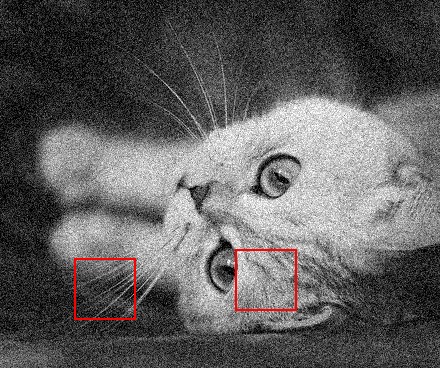} &
		\includegraphics[width = 0.33\textwidth]{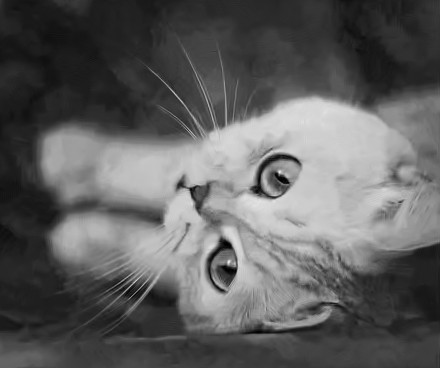} \\
		
	\end{tabular}  
	\begin{tabular}{c@{\hskip 0.005\textwidth}c@{\hskip 0.005\textwidth}c@{\hskip 0.005\textwidth}c@{\hskip 0.005\textwidth}c@{\hskip 0.005\textwidth}c}
		
		\includegraphics[width = 0.1625\textwidth]{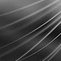} &
		\includegraphics[width = 0.1625\textwidth]{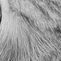} &
		\includegraphics[width = 0.1625\textwidth]{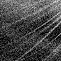} &
		\includegraphics[width = 0.1625\textwidth]{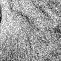} &
		\includegraphics[width = 0.1625\textwidth]{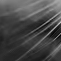} &
		\includegraphics[width = 0.1625\textwidth]{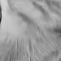} \\
		
		\multicolumn{2}{c}{Ground truth}&
		\multicolumn{2}{c}{Noisy image}&
		\multicolumn{2}{c}{BM3D}\\   
		
		\multicolumn{2}{c}{$$}&
		\multicolumn{2}{c}{$$}&
		\multicolumn{2}{c}{$31.80dB$}\\   
		
		\smallskip
	\end{tabular}  
	
	\begin{tabular}{c@{\hskip 0.005\textwidth}c@{\hskip 0.005\textwidth}c}
		
		\includegraphics[width = 0.33\textwidth]{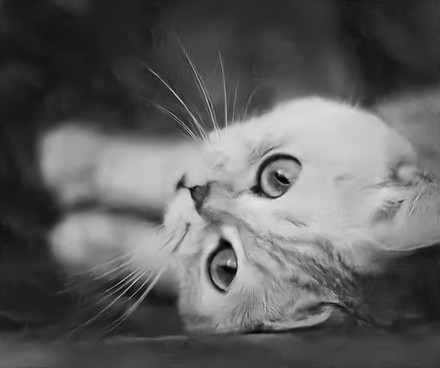} &
		\includegraphics[width = 0.33\textwidth]{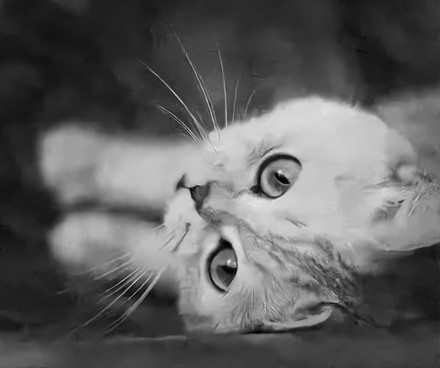} &
		\includegraphics[width = 0.33\textwidth]{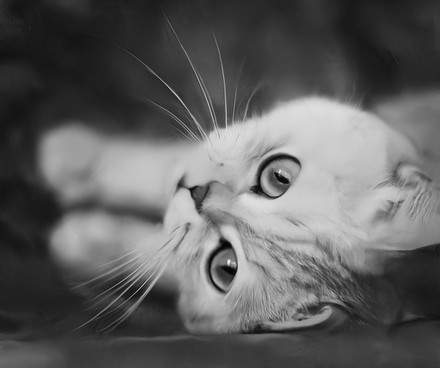} \\ 
	\end{tabular} 
	\begin{tabular}{c@{\hskip 0.005\textwidth}c@{\hskip 0.005\textwidth}c@{\hskip 0.005\textwidth}c@{\hskip 0.005\textwidth}c@{\hskip 0.005\textwidth}c}
		
		\includegraphics[width = 0.1625\textwidth]{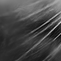} &
		\includegraphics[width = 0.1625\textwidth]{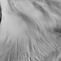} &
		\includegraphics[width = 0.1625\textwidth]{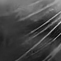} &
		\includegraphics[width = 0.1625\textwidth]{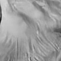} &
		\includegraphics[width = 0.1625\textwidth]{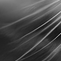} &
		\includegraphics[width = 0.1625\textwidth]{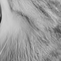} \\  
		
		\multicolumn{2}{c}{MLP}&
		\multicolumn{2}{c}{TNRD}&
		\multicolumn{2}{c}{\textit{Pet}-specific DenoiseNet}\\  
		
		\multicolumn{2}{c}{$29.97dB$ }&
		\multicolumn{2}{c}{$31.28dB$ }&
		\multicolumn{2}{c}{$32.37dB$}\\   
		
	\end{tabular}  
\end{figure*}
\bigskip 

\begin{figure*}[]  
	\centering 
	\begin{tabular}{c@{\hskip 0.005\textwidth}c@{\hskip 0.005\textwidth}c}
		
		\includegraphics[width = 0.33\textwidth]{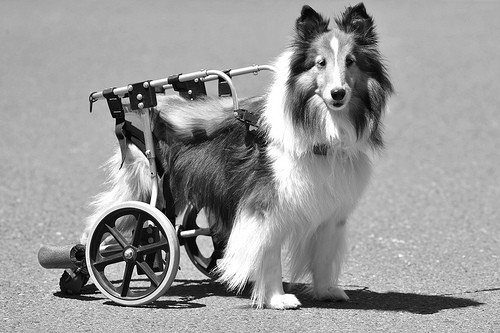} &
		\includegraphics[width = 0.33\textwidth]{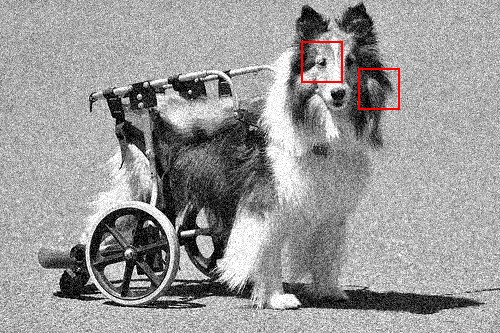} &
		\includegraphics[width = 0.33\textwidth]{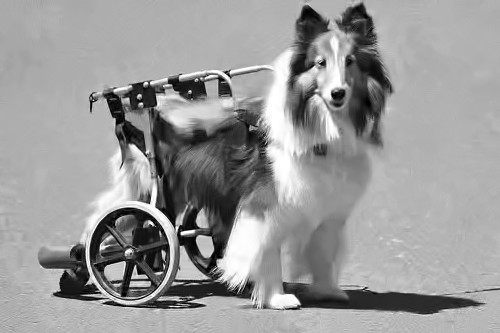} \\
		
	\end{tabular}  
	\begin{tabular}{c@{\hskip 0.005\textwidth}c@{\hskip 0.005\textwidth}c@{\hskip 0.005\textwidth}c@{\hskip 0.005\textwidth}c@{\hskip 0.005\textwidth}c}
		
		\includegraphics[width = 0.1625\textwidth]{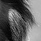} &
		\includegraphics[width = 0.1625\textwidth]{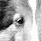} &
		\includegraphics[width = 0.1625\textwidth]{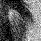} &
		\includegraphics[width = 0.1625\textwidth]{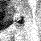} &
		\includegraphics[width = 0.1625\textwidth]{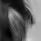} &
		\includegraphics[width = 0.1625\textwidth]{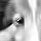} \\
		
		\multicolumn{2}{c}{Ground truth}&
		\multicolumn{2}{c}{Noisy image}&
		\multicolumn{2}{c}{BM3D}\\
		
		\multicolumn{2}{c}{$$}&
		\multicolumn{2}{c}{$$}&
		\multicolumn{2}{c}{$27.82dB$}\\  
		
		\smallskip      
	\end{tabular}         
	\begin{tabular}{c@{\hskip 0.005\textwidth}c@{\hskip 0.005\textwidth}c}
		
		\includegraphics[width = 0.33\textwidth]{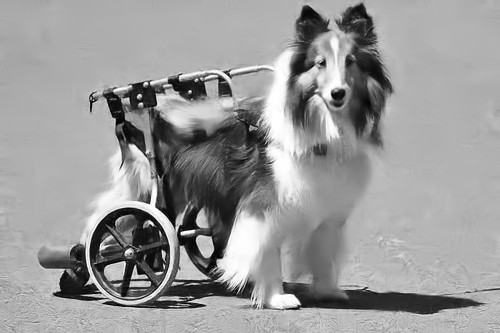} &
		\includegraphics[width = 0.33\textwidth]{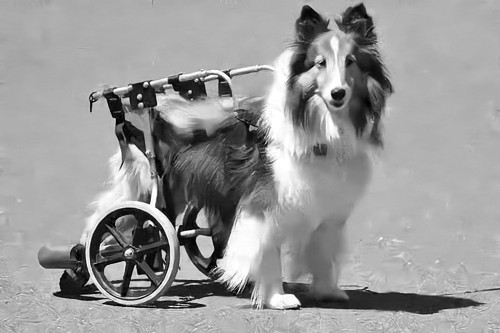} &
		\includegraphics[width = 0.33\textwidth]{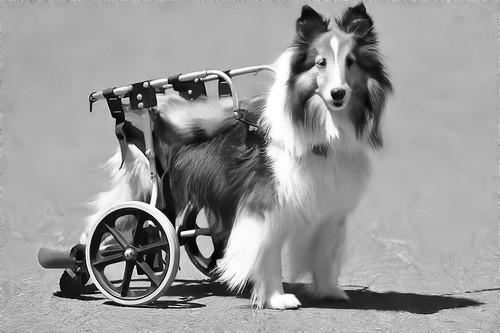} \\
		
	\end{tabular}  
	\begin{tabular}{c@{\hskip 0.005\textwidth}c@{\hskip 0.005\textwidth}c@{\hskip 0.005\textwidth}c@{\hskip 0.005\textwidth}c@{\hskip 0.005\textwidth}c}
		
		\includegraphics[width = 0.1625\textwidth]{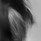} &
		\includegraphics[width = 0.1625\textwidth]{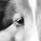} &
		\includegraphics[width = 0.1625\textwidth]{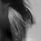} &
		\includegraphics[width = 0.1625\textwidth]{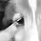} &
		\includegraphics[width = 0.1625\textwidth]{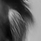} &
		\includegraphics[width = 0.1625\textwidth]{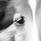} \\
		
		\multicolumn{2}{c}{MLP}&
		\multicolumn{2}{c}{TNRD}&
		\multicolumn{2}{c}{\textit{Pet}-specific DenoiseNet}\\
		
		\multicolumn{2}{c}{$28.23dB$}&
		\multicolumn{2}{c}{$28.24dB$}&
		\multicolumn{2}{c}{$28.69dB$}\\   
		
	\end{tabular}  
\end{figure*}
\bigskip 
\begin{figure*}[]  
	\centering     
	\begin{tabular}{c@{\hskip 0.005\textwidth}c@{\hskip 0.005\textwidth}c}
		
		\includegraphics[width = 0.3\textwidth]{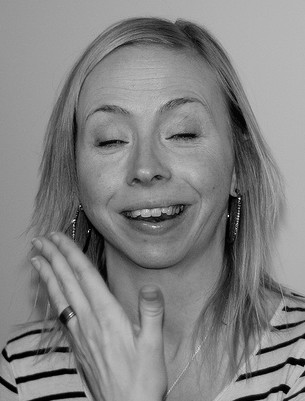} &
		\includegraphics[width = 0.3\textwidth]{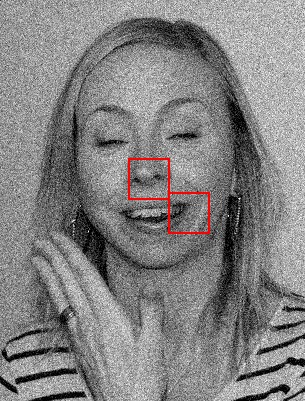} &
		\includegraphics[width = 0.3\textwidth]{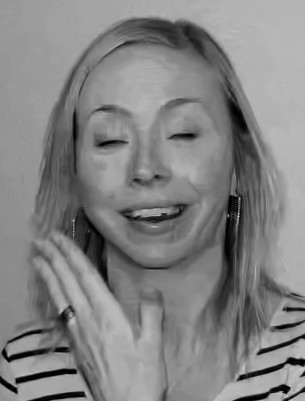} \\
		
	\end{tabular}  
	\begin{tabular}{c@{\hskip 0.005\textwidth}c@{\hskip 0.005\textwidth}c@{\hskip 0.005\textwidth}c@{\hskip 0.005\textwidth}c@{\hskip 0.005\textwidth}c}
		
		\includegraphics[width = 0.1475\textwidth]{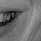} &
		\includegraphics[width = 0.1475\textwidth]{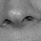} &
		\includegraphics[width = 0.1475\textwidth]{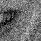} &
		\includegraphics[width = 0.1475\textwidth]{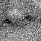} &
		\includegraphics[width = 0.1475\textwidth]{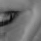} &
		\includegraphics[width = 0.1475\textwidth]{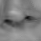} \\
		
		\multicolumn{2}{c}{Ground truth}&
		\multicolumn{2}{c}{Noisy image}&
		\multicolumn{2}{c}{BM3D}\\
		
		\multicolumn{2}{c}{$$}&
		\multicolumn{2}{c}{$$}&
		\multicolumn{2}{c}{$31.63dB$}\\ 
		
		\smallskip
	\end{tabular} 
	\begin{tabular}{c@{\hskip 0.005\textwidth}c@{\hskip 0.005\textwidth}c}
		
		\includegraphics[width = 0.3\textwidth]{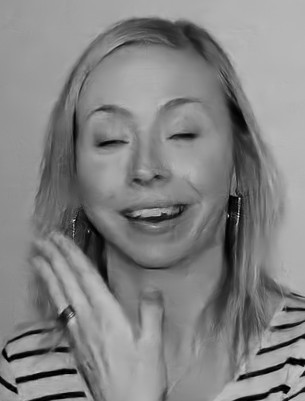} &
		\includegraphics[width = 0.3\textwidth]{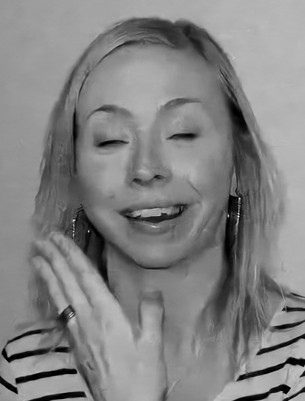} &
		\includegraphics[width = 0.3\textwidth]{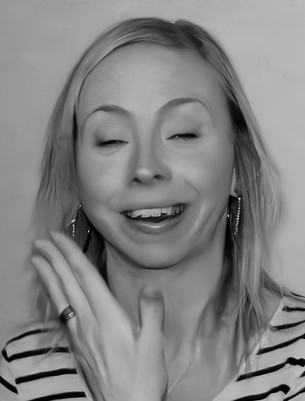} \\
		
	\end{tabular}  
	\begin{tabular}{c@{\hskip 0.005\textwidth}c@{\hskip 0.005\textwidth}c@{\hskip 0.005\textwidth}c@{\hskip 0.005\textwidth}c@{\hskip 0.005\textwidth}c}
		
		\includegraphics[width = 0.1475\textwidth]{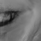} &
		\includegraphics[width = 0.1475\textwidth]{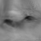} &
		\includegraphics[width = 0.1475\textwidth]{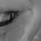} &
		\includegraphics[width = 0.1475\textwidth]{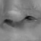} &
		\includegraphics[width = 0.1475\textwidth]{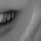} &
		\includegraphics[width = 0.1475\textwidth]{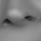} \\
		
		\multicolumn{2}{c}{MLP}&
		\multicolumn{2}{c}{TNRD}&
		\multicolumn{2}{c}{\textit{Face}-specific DenoiseNet}\\
		
		\multicolumn{2}{c}{$31.68dB$} &
		\multicolumn{2}{c}{$31.73dB$} &
		\multicolumn{2}{c}{$32.46dB$} \\ 
		
	\end{tabular} 
\end{figure*}
\bigskip

\begin{figure*}[]
	\centering   
	\begin{tabular}{c@{\hskip 0.005\textwidth}c@{\hskip 0.005\textwidth}c}
		
		\includegraphics[width = 0.3\textwidth]{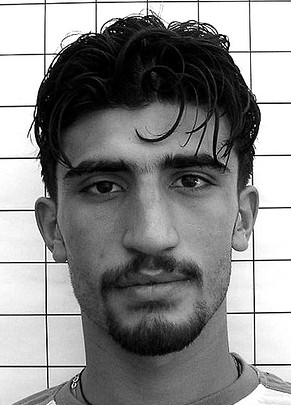} &
		\includegraphics[width = 0.3\textwidth]{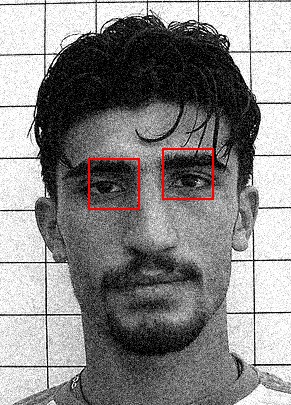} &
		\includegraphics[width = 0.3\textwidth]{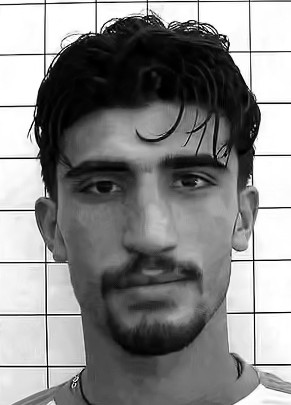} \\
		
	\end{tabular}  
	\begin{tabular}{c@{\hskip 0.005\textwidth}c@{\hskip 0.005\textwidth}c@{\hskip 0.005\textwidth}c@{\hskip 0.005\textwidth}c@{\hskip 0.005\textwidth}c}
		
		\includegraphics[width = 0.1475\textwidth]{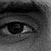} &
		\includegraphics[width = 0.1475\textwidth]{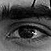} &
		\includegraphics[width = 0.1475\textwidth]{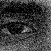} &
		\includegraphics[width = 0.1475\textwidth]{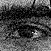} &
		\includegraphics[width = 0.1475\textwidth]{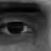} &
		\includegraphics[width = 0.1475\textwidth]{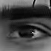} \\
		
		\multicolumn{2}{c}{Ground truth}&
		\multicolumn{2}{c}{Noisy image}&
		\multicolumn{2}{c}{BM3D}\\
		
		\multicolumn{2}{c}{$$}&
		\multicolumn{2}{c}{$$}&
		\multicolumn{2}{c}{$30.23dB$}\\ 
		
		\smallskip        
	\end{tabular} 
	\begin{tabular}{c@{\hskip 0.005\textwidth}c@{\hskip 0.005\textwidth}c}
		
		\includegraphics[width = 0.3\textwidth]{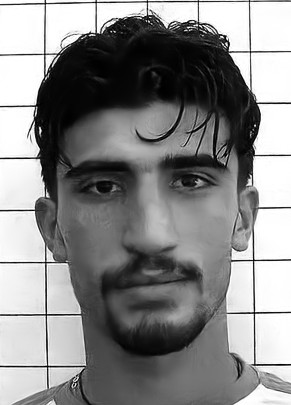} &
		\includegraphics[width = 0.3\textwidth]{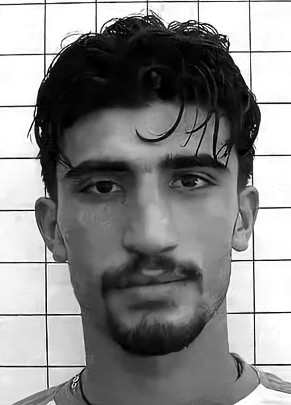} &
		\includegraphics[width = 0.3\textwidth]{./n09618957_19032_bm3d.jpg} \\
		
	\end{tabular}  
	\begin{tabular}{c@{\hskip 0.005\textwidth}c@{\hskip 0.005\textwidth}c@{\hskip 0.005\textwidth}c@{\hskip 0.005\textwidth}c@{\hskip 0.005\textwidth}c}
		
		\includegraphics[width = 0.1475\textwidth]{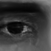} &
		\includegraphics[width = 0.1475\textwidth]{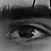} &
		\includegraphics[width = 0.1475\textwidth]{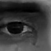} &
		\includegraphics[width = 0.1475\textwidth]{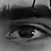} &
		\includegraphics[width = 0.1475\textwidth]{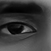} &
		\includegraphics[width = 0.1475\textwidth]{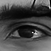} \\
		
		\multicolumn{2}{c}{MLP}&
		\multicolumn{2}{c}{TNRD}&
		\multicolumn{2}{c}{\textit{Face}-specific DenoiseNet}\\
		
		\multicolumn{2}{c}{$29.99dB$}&
		\multicolumn{2}{c}{$30.04dB$}&
		\multicolumn{2}{c}{$30.99dB$}\\ 
		
	\end{tabular} 
	
\end{figure*}
\bigskip 
\begin{figure*}[]  
	\centering     
	\begin{tabular}{c@{\hskip 0.005\textwidth}c@{\hskip 0.005\textwidth}c}
		
		\includegraphics[width = 0.33\textwidth]{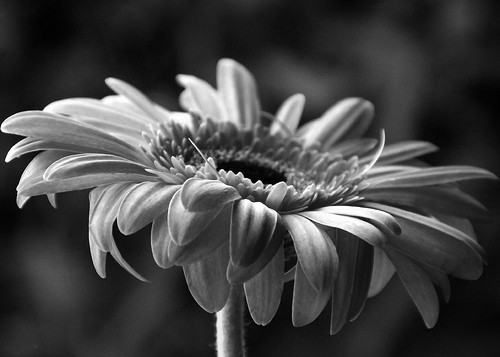} &
		\includegraphics[width = 0.33\textwidth]{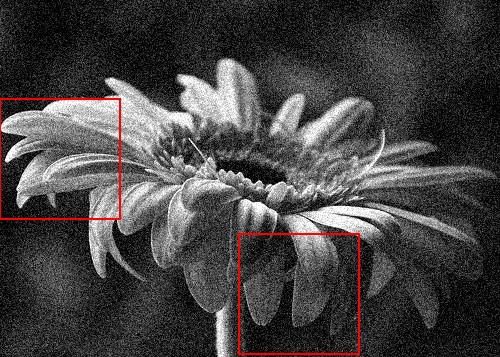} &
		\includegraphics[width = 0.33\textwidth]{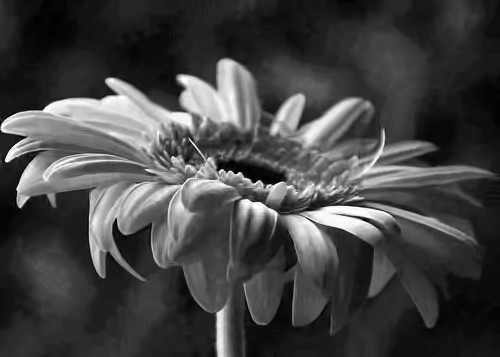} \\
		
	\end{tabular}  
	\begin{tabular}{c@{\hskip 0.005\textwidth}c@{\hskip 0.005\textwidth}c@{\hskip 0.005\textwidth}c@{\hskip 0.005\textwidth}c@{\hskip 0.005\textwidth}c}
		
		\includegraphics[width = 0.1625\textwidth]{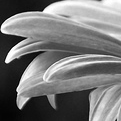} &
		\includegraphics[width = 0.1625\textwidth]{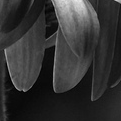} &
		\includegraphics[width = 0.1625\textwidth]{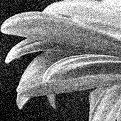} &
		\includegraphics[width = 0.1625\textwidth]{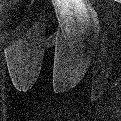} &
		\includegraphics[width = 0.1625\textwidth]{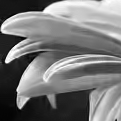} &
		\includegraphics[width = 0.1625\textwidth]{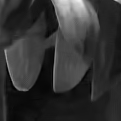} \\
		
		\multicolumn{2}{c}{Ground truth}&
		\multicolumn{2}{c}{Noisy image}&
		\multicolumn{2}{c}{BM3D}\\
		
		\multicolumn{2}{c}{$$}&
		\multicolumn{2}{c}{$$}&
		\multicolumn{2}{c}{$31.66dB$}\\ 
		
		\smallskip        
	\end{tabular}  
	
	\begin{tabular}{c@{\hskip 0.005\textwidth}c@{\hskip 0.005\textwidth}c}
		
		\includegraphics[width = 0.33\textwidth]{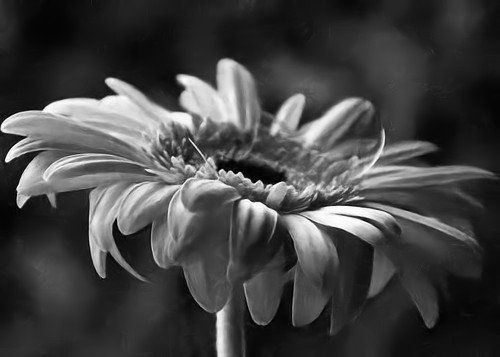} &
		\includegraphics[width = 0.33\textwidth]{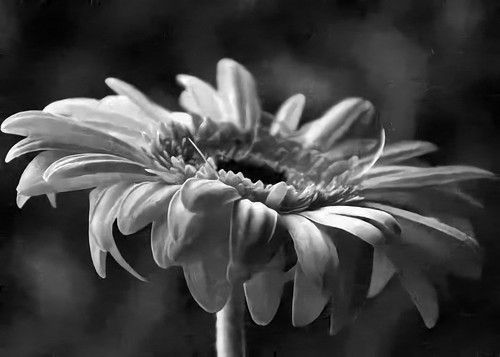} &
		\includegraphics[width = 0.33\textwidth]{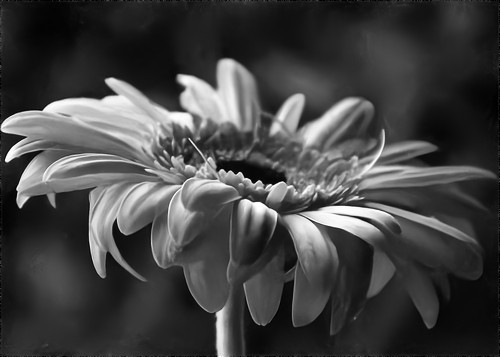} \\
		
	\end{tabular}  
	\begin{tabular}{c@{\hskip 0.005\textwidth}c@{\hskip 0.005\textwidth}c@{\hskip 0.005\textwidth}c@{\hskip 0.005\textwidth}c@{\hskip 0.005\textwidth}c}
		
		\includegraphics[width = 0.1625\textwidth]{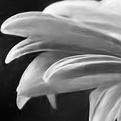} &
		\includegraphics[width = 0.1625\textwidth]{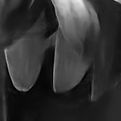} &
		\includegraphics[width = 0.1625\textwidth]{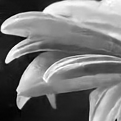} &
		\includegraphics[width = 0.1625\textwidth]{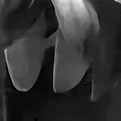} &
		\includegraphics[width = 0.1625\textwidth]{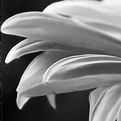} &
		\includegraphics[width = 0.1625\textwidth]{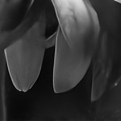} \\
		
		\multicolumn{2}{c}{MLP}&
		\multicolumn{2}{c}{TNRD}&
		\multicolumn{2}{c}{\textit{Flower}-specific DenoiseNet}\\
		
		\multicolumn{2}{c}{$31.97dB$}&
		\multicolumn{2}{c}{$31.86dB$}&
		\multicolumn{2}{c}{$32.75dB$}\\

	\end{tabular} 
\end{figure*}
\bigskip 
\begin{figure*}[]  
	\centering     
	\begin{tabular}{c@{\hskip 0.005\textwidth}c@{\hskip 0.005\textwidth}c}
		
		\includegraphics[width = 0.33\textwidth]{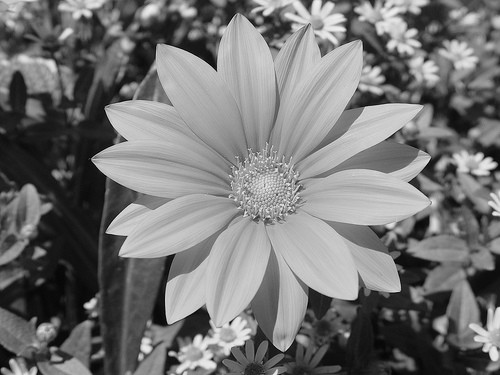} &
		\includegraphics[width = 0.33\textwidth]{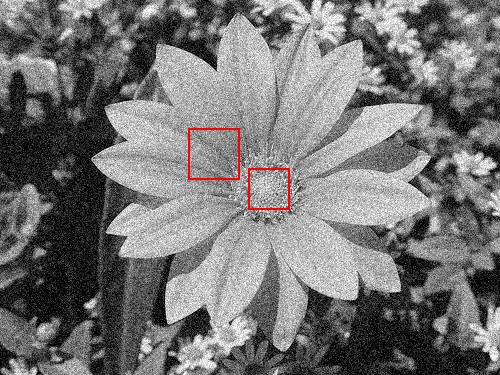} &
		\includegraphics[width = 0.33\textwidth]{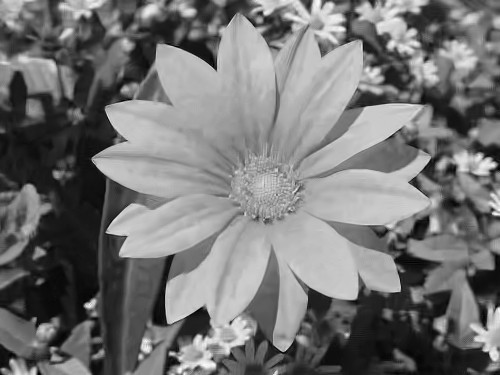} \\
		
	\end{tabular}  
	\begin{tabular}{c@{\hskip 0.005\textwidth}c@{\hskip 0.005\textwidth}c@{\hskip 0.005\textwidth}c@{\hskip 0.005\textwidth}c@{\hskip 0.005\textwidth}c}
		
		\includegraphics[width = 0.1625\textwidth]{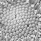} &
		\includegraphics[width = 0.1625\textwidth]{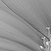} &
		\includegraphics[width = 0.1625\textwidth]{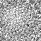} &
		\includegraphics[width = 0.1625\textwidth]{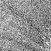} &
		\includegraphics[width = 0.1625\textwidth]{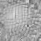} &
		\includegraphics[width = 0.1625\textwidth]{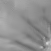} \\
		
		\multicolumn{2}{c}{Ground truth}&
		\multicolumn{2}{c}{Noisy image}&
		\multicolumn{2}{c}{BM3D} \\
		
		\multicolumn{2}{c}{$$}&
		\multicolumn{2}{c}{$$}&
		\multicolumn{2}{c}{$30.39dB$}\\  
		
		\smallskip
	\end{tabular}    
	\begin{tabular}{c@{\hskip 0.005\textwidth}c@{\hskip 0.005\textwidth}c}
		
		\includegraphics[width = 0.33\textwidth]{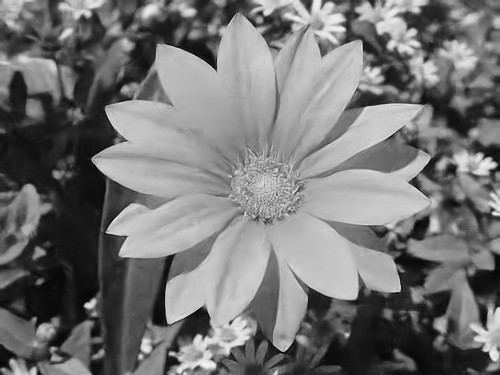} &
		\includegraphics[width = 0.33\textwidth]{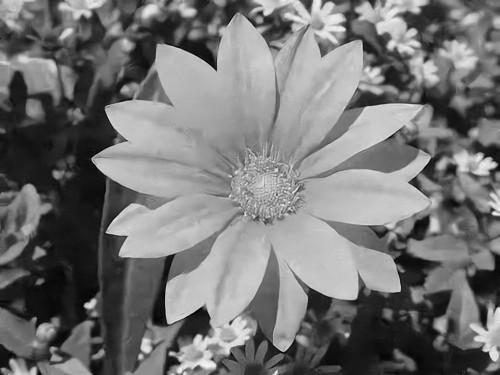} &
		\includegraphics[width = 0.33\textwidth]{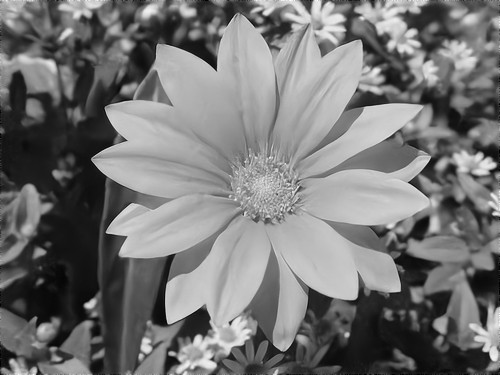} \\
		
	\end{tabular}  
	\begin{tabular}{c@{\hskip 0.005\textwidth}c@{\hskip 0.005\textwidth}c@{\hskip 0.005\textwidth}c@{\hskip 0.005\textwidth}c@{\hskip 0.005\textwidth}c}
		
		\includegraphics[width = 0.1625\textwidth]{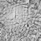} &
		\includegraphics[width = 0.1625\textwidth]{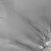} &
		\includegraphics[width = 0.1625\textwidth]{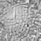} &
		\includegraphics[width = 0.1625\textwidth]{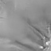} &
		\includegraphics[width = 0.1625\textwidth]{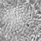} &
		\includegraphics[width = 0.1625\textwidth]{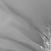} \\
		
		\multicolumn{2}{c}{MLP}&
		\multicolumn{2}{c}{TNRD}&
		\multicolumn{2}{c}{\textit{Flower}-specific DenoiseNet} \\
		
		\multicolumn{2}{c}{$30.64dB$}&
		\multicolumn{2}{c}{$30.67dB$}&
		\multicolumn{2}{c}{$31.37dB$}\\
		
	\end{tabular} 
\end{figure*}
\bigskip

\begin{figure*}[]  
	\centering     
	\begin{tabular}{c@{\hskip 0.005\textwidth}c@{\hskip 0.005\textwidth}c}
		
		\includegraphics[width = 0.33\textwidth]{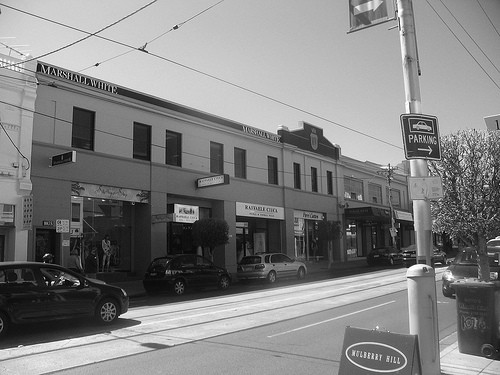} &
		\includegraphics[width = 0.33\textwidth]{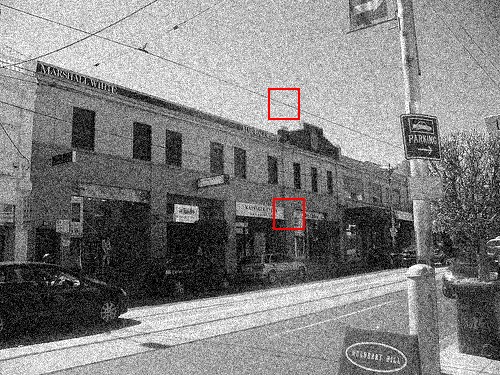} &
		\includegraphics[width = 0.33\textwidth]{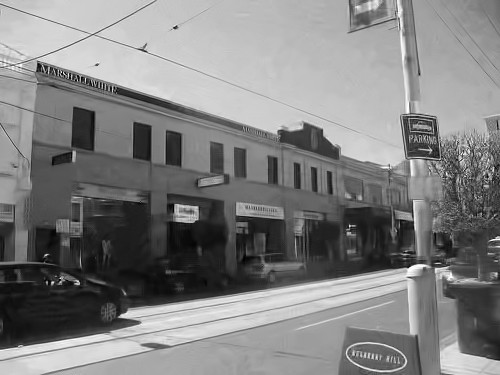} \\
		
	\end{tabular}  
	\begin{tabular}{c@{\hskip 0.005\textwidth}c@{\hskip 0.005\textwidth}c@{\hskip 0.005\textwidth}c@{\hskip 0.005\textwidth}c@{\hskip 0.005\textwidth}c}
		
		\includegraphics[width = 0.1625\textwidth]{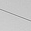} &
		\includegraphics[width = 0.1625\textwidth]{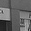} &
		\includegraphics[width = 0.1625\textwidth]{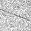} &
		\includegraphics[width = 0.1625\textwidth]{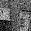} &
		\includegraphics[width = 0.1625\textwidth]{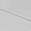} &
		\includegraphics[width = 0.1625\textwidth]{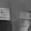} \\
		
		\multicolumn{2}{c}{Ground truth}&
		\multicolumn{2}{c}{Noisy image}&
		\multicolumn{2}{c}{BM3D} \\
		
		\multicolumn{2}{c}{$$}&
		\multicolumn{2}{c}{$$}&
		\multicolumn{2}{c}{$28.67dB$}\\  
		
		\smallskip        
	\end{tabular}     
	\begin{tabular}{c@{\hskip 0.005\textwidth}c@{\hskip 0.005\textwidth}c}
		
		\includegraphics[width = 0.33\textwidth]{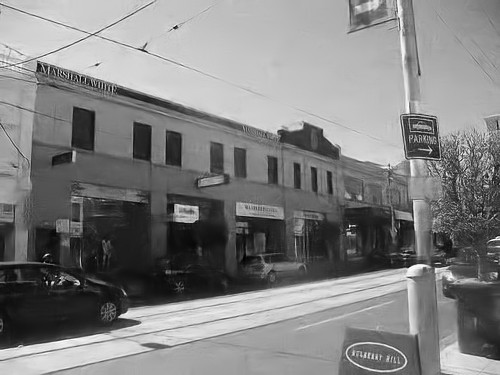} &
		\includegraphics[width = 0.33\textwidth]{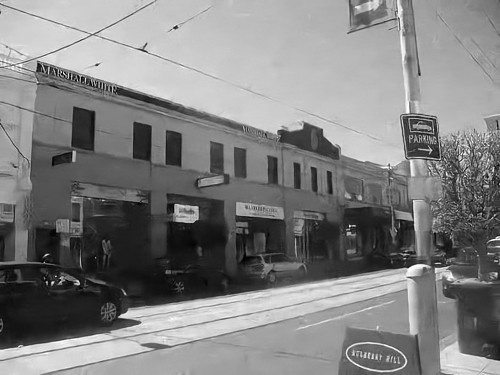} &
		\includegraphics[width = 0.33\textwidth]{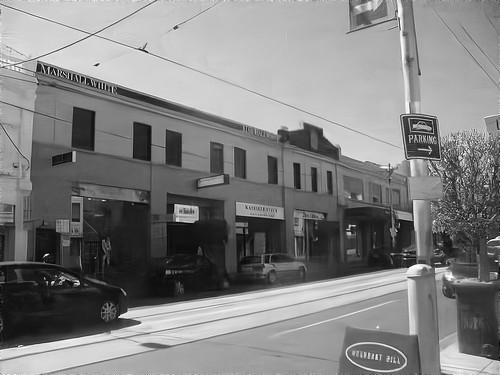} \\
		
	\end{tabular}  
	\begin{tabular}{c@{\hskip 0.005\textwidth}c@{\hskip 0.005\textwidth}c@{\hskip 0.005\textwidth}c@{\hskip 0.005\textwidth}c@{\hskip 0.005\textwidth}c}
		
		\includegraphics[width = 0.1625\textwidth]{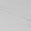} &
		\includegraphics[width = 0.1625\textwidth]{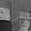} &
		\includegraphics[width = 0.1625\textwidth]{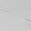} &
		\includegraphics[width = 0.1625\textwidth]{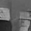} &
		\includegraphics[width = 0.1625\textwidth]{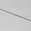} &
		\includegraphics[width = 0.1625\textwidth]{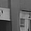} \\
		
		\multicolumn{2}{c}{MLP}&
		\multicolumn{2}{c}{TNRD}&
		\multicolumn{2}{c}{\textit{Street}-specific DenoiseNet} \\
		
		\multicolumn{2}{c}{$28.71dB$}&
		\multicolumn{2}{c}{$28.76dB$}&
		\multicolumn{2}{c}{$29.76dB$}\\  
		
	\end{tabular}   
\end{figure*}
\bigskip 
\begin{figure*}[]  
	\centering     
	\begin{tabular}{c@{\hskip 0.005\textwidth}c@{\hskip 0.005\textwidth}c}
		
		\includegraphics[width = 0.33\textwidth]{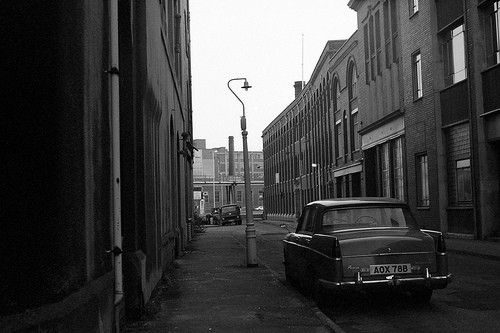} &
		\includegraphics[width = 0.33\textwidth]{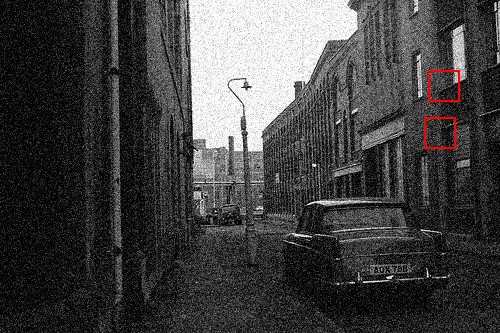} &
		\includegraphics[width = 0.33\textwidth]{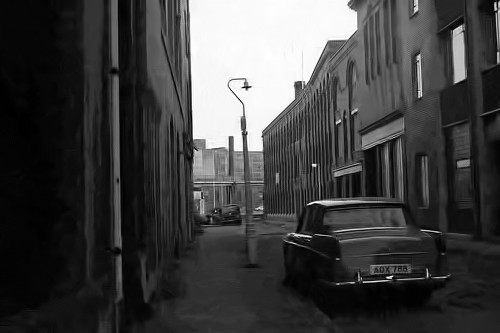} \\
		
	\end{tabular}  
	\begin{tabular}{c@{\hskip 0.005\textwidth}c@{\hskip 0.005\textwidth}c@{\hskip 0.005\textwidth}c@{\hskip 0.005\textwidth}c@{\hskip 0.005\textwidth}c}
		
		\includegraphics[width = 0.1625\textwidth]{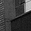} &
		\includegraphics[width = 0.1625\textwidth]{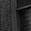} &
		\includegraphics[width = 0.1625\textwidth]{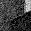} &
		\includegraphics[width = 0.1625\textwidth]{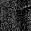} &
		\includegraphics[width = 0.1625\textwidth]{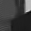} &
		\includegraphics[width = 0.1625\textwidth]{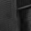} \\
		
		\multicolumn{2}{c}{Ground truth}&
		\multicolumn{2}{c}{Noisy image}&
		\multicolumn{2}{c}{BM3D} \\
		
		\multicolumn{2}{c}{$$}&
		\multicolumn{2}{c}{$$}&
		\multicolumn{2}{c}{$29.34dB$}\\  
		
		\smallskip        
	\end{tabular}  
	\begin{tabular}{c@{\hskip 0.005\textwidth}c@{\hskip 0.005\textwidth}c}
		
		\includegraphics[width = 0.33\textwidth]{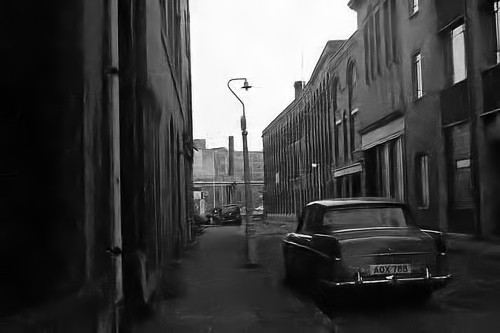} &
		\includegraphics[width = 0.33\textwidth]{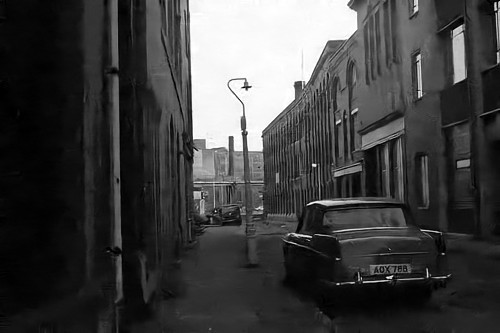} &
		\includegraphics[width = 0.33\textwidth]{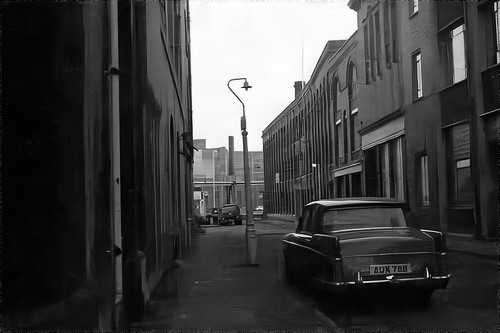} \\
		
	\end{tabular}  
	\begin{tabular}{c@{\hskip 0.005\textwidth}c@{\hskip 0.005\textwidth}c@{\hskip 0.005\textwidth}c@{\hskip 0.005\textwidth}c@{\hskip 0.005\textwidth}c}
		
		\includegraphics[width = 0.1625\textwidth]{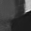} &
		\includegraphics[width = 0.1625\textwidth]{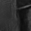} &
		\includegraphics[width = 0.1625\textwidth]{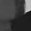} &
		\includegraphics[width = 0.1625\textwidth]{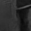} &
		\includegraphics[width = 0.1625\textwidth]{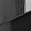} &
		\includegraphics[width = 0.1625\textwidth]{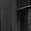} \\
		
		\multicolumn{2}{c}{MLP}&
		\multicolumn{2}{c}{TNRD}&
		\multicolumn{2}{c}{\textit{Street}-specific DenoiseNet} \\
		
		\multicolumn{2}{c}{$29.33dB$}&
		\multicolumn{2}{c}{$29.29dB$}&
		\multicolumn{2}{c}{$30.13dB$}\\  
		
		\smallskip        
	\end{tabular}      
\end{figure*}
\bigskip 
\begin{figure*}[]  
	\centering     
	\begin{tabular}{c@{\hskip 0.005\textwidth}c@{\hskip 0.005\textwidth}c}
		
		\includegraphics[width = 0.33\textwidth]{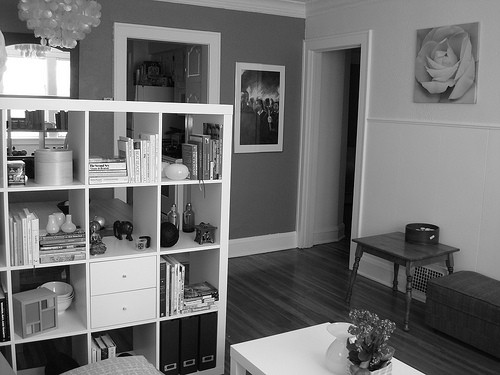} &
		\includegraphics[width = 0.33\textwidth]{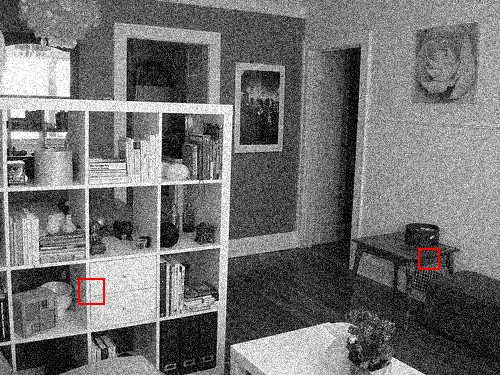} &
		\includegraphics[width = 0.33\textwidth]{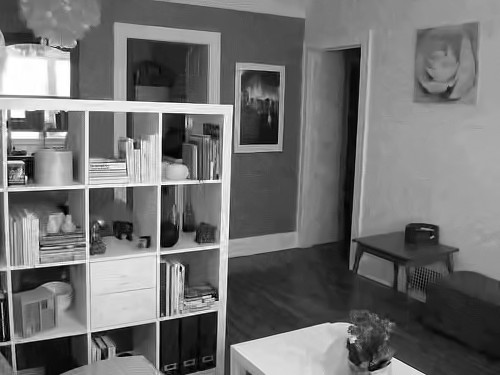} \\
		
	\end{tabular}  
	\begin{tabular}{c@{\hskip 0.005\textwidth}c@{\hskip 0.005\textwidth}c@{\hskip 0.005\textwidth}c@{\hskip 0.005\textwidth}c@{\hskip 0.005\textwidth}c}
		
		\includegraphics[width = 0.1625\textwidth]{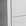} &
		\includegraphics[width = 0.1625\textwidth]{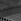} &
		\includegraphics[width = 0.1625\textwidth]{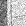} &
		\includegraphics[width = 0.1625\textwidth]{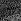} &
		\includegraphics[width = 0.1625\textwidth]{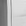} &
		\includegraphics[width = 0.1625\textwidth]{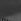}  \\
		
		\multicolumn{2}{c}{Ground truth}&
		\multicolumn{2}{c}{Noisy image}&
		\multicolumn{2}{c}{BM3D} \\
		
		\multicolumn{2}{c}{$$}&
		\multicolumn{2}{c}{$$}&
		\multicolumn{2}{c}{$29.94dB$}\\ 
		
		\smallskip       
	\end{tabular} 
	\begin{tabular}{c@{\hskip 0.005\textwidth}c@{\hskip 0.005\textwidth}c}
		
		\includegraphics[width = 0.33\textwidth]{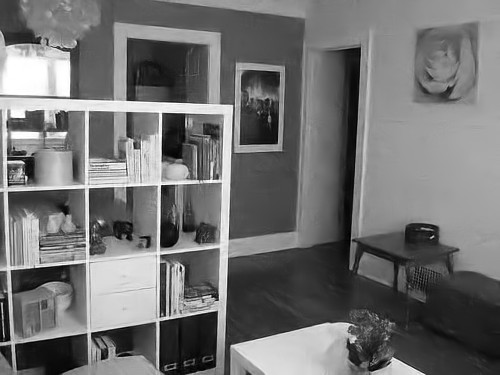} &
		\includegraphics[width = 0.33\textwidth]{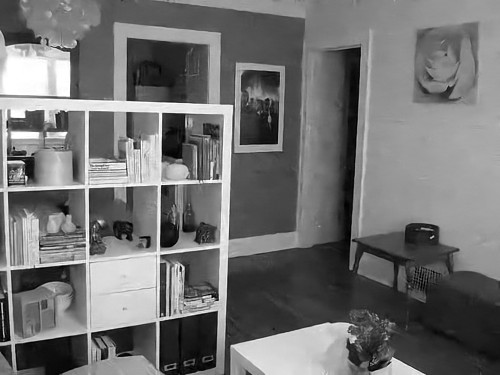} &
		\includegraphics[width = 0.33\textwidth]{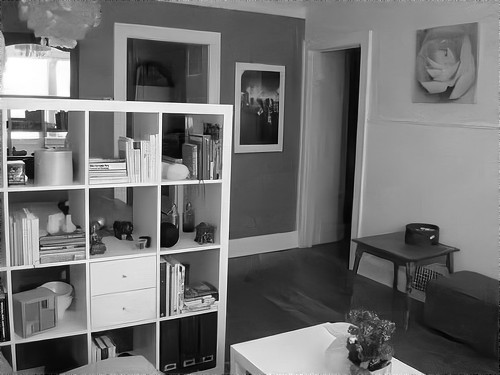} \\
		
	\end{tabular}  
	\begin{tabular}{c@{\hskip 0.005\textwidth}c@{\hskip 0.005\textwidth}c@{\hskip 0.005\textwidth}c@{\hskip 0.005\textwidth}c@{\hskip 0.005\textwidth}c}
		
		\includegraphics[width = 0.1625\textwidth]{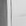} &
		\includegraphics[width = 0.1625\textwidth]{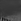} &
		\includegraphics[width = 0.1625\textwidth]{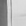} &
		\includegraphics[width = 0.1625\textwidth]{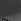} &
		\includegraphics[width = 0.1625\textwidth]{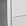} &
		\includegraphics[width = 0.1625\textwidth]{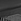}  \\
		
		\multicolumn{2}{c}{MLP}&
		\multicolumn{2}{c}{TNRD}&
		\multicolumn{2}{c}{\textit{Living room}-specific DenoiseNet} \\
		
		\multicolumn{2}{c}{$29.94dB$}&
		\multicolumn{2}{c}{$30.13dB$}&
		\multicolumn{2}{c}{$31.13dB$}\\
		
		\smallskip       
	\end{tabular}        
\end{figure*}
\bigskip

\begin{figure*}[]  
	\centering     
	\begin{tabular}{c@{\hskip 0.005\textwidth}c@{\hskip 0.005\textwidth}c}
		
		\includegraphics[width = 0.33\textwidth]{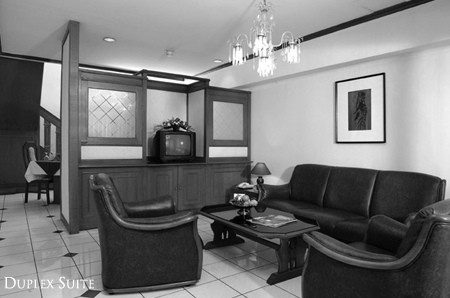} &
		\includegraphics[width = 0.33\textwidth]{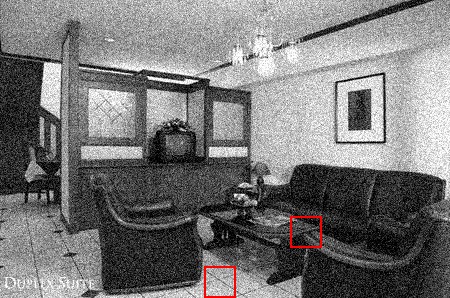} &
		\includegraphics[width = 0.33\textwidth]{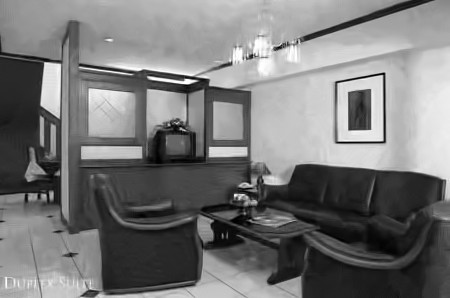}  \\
		
	\end{tabular}  
	\begin{tabular}{c@{\hskip 0.005\textwidth}c@{\hskip 0.005\textwidth}c@{\hskip 0.005\textwidth}c@{\hskip 0.005\textwidth}c@{\hskip 0.005\textwidth}c}
		
		\includegraphics[width = 0.1625\textwidth]{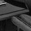} &
		\includegraphics[width = 0.1625\textwidth]{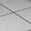} &
		\includegraphics[width = 0.1625\textwidth]{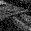} &
		\includegraphics[width = 0.1625\textwidth]{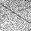} &
		\includegraphics[width = 0.1625\textwidth]{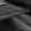} &
		\includegraphics[width = 0.1625\textwidth]{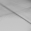} \\
		
		\multicolumn{2}{c}{Ground truth}&
		\multicolumn{2}{c}{Noisy image}&
		\multicolumn{2}{c}{BM3D} \\
		
		\multicolumn{2}{c}{}&
		\multicolumn{2}{c}{}&
		\multicolumn{2}{c}{$30.99dB$}\\
		
		\smallskip
		
	\end{tabular}    
	\begin{tabular}{c@{\hskip 0.005\textwidth}c@{\hskip 0.005\textwidth}c}
		
		\includegraphics[width = 0.33\textwidth]{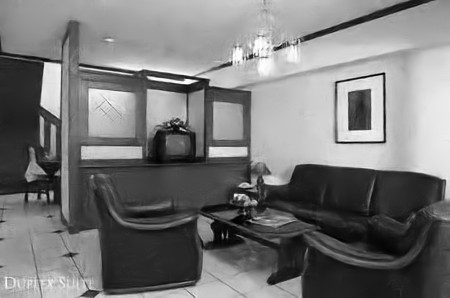} &
		\includegraphics[width = 0.33\textwidth]{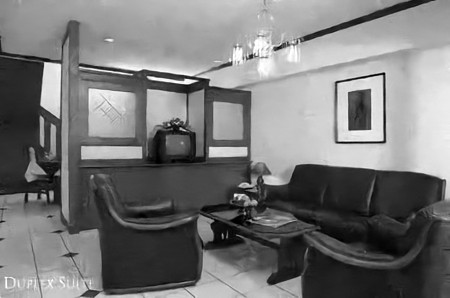} &
		\includegraphics[width = 0.33\textwidth]{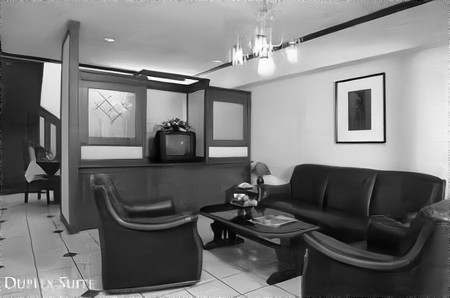}  \\
		
	\end{tabular}  
	\begin{tabular}{c@{\hskip 0.005\textwidth}c@{\hskip 0.005\textwidth}c@{\hskip 0.005\textwidth}c@{\hskip 0.005\textwidth}c@{\hskip 0.005\textwidth}c}
		
		\includegraphics[width = 0.1625\textwidth]{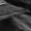} &
		\includegraphics[width = 0.1625\textwidth]{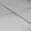} &
		\includegraphics[width = 0.1625\textwidth]{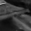} &
		\includegraphics[width = 0.1625\textwidth]{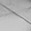} &
		\includegraphics[width = 0.1625\textwidth]{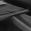} &
		\includegraphics[width = 0.1625\textwidth]{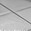} \\
		
		\multicolumn{2}{c}{MLP}&
		\multicolumn{2}{c}{TNRD}&
		\multicolumn{2}{c}{\textit{Living room}-specific DenoiseNet} \\
		
		\multicolumn{2}{c}{$31.03dB$}& 
		\multicolumn{2}{c}{$31.13dB$}&
		\multicolumn{2}{c}{$31.99dB$}\\         
		
	\end{tabular}      
\end{figure*}
\bigskip 

\begin{figure*}[]  
	\centering    
	\begin{tabular}{c@{\hskip 0.005\textwidth}c@{\hskip 0.005\textwidth}c}
		
		\includegraphics[width = 0.33\textwidth]{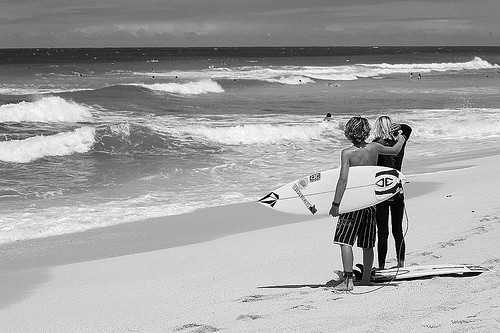} &
		\includegraphics[width = 0.33\textwidth]{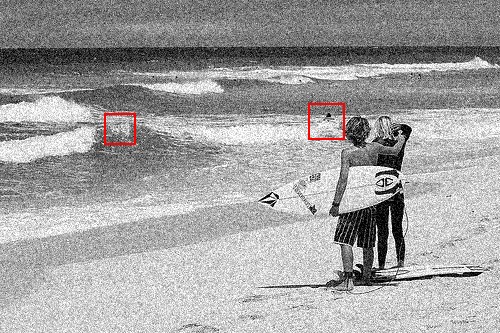} &
		\includegraphics[width = 0.33\textwidth]{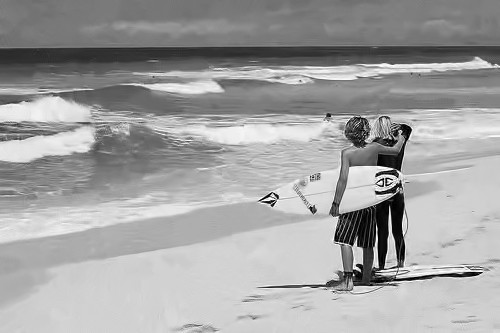} \\
		
	\end{tabular}  
	\begin{tabular}{c@{\hskip 0.005\textwidth}c@{\hskip 0.005\textwidth}c@{\hskip 0.005\textwidth}c@{\hskip 0.005\textwidth}c@{\hskip 0.005\textwidth}c}
		
		\includegraphics[width = 0.1625\textwidth]{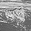} &
		\includegraphics[width = 0.1625\textwidth]{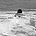} &
		\includegraphics[width = 0.1625\textwidth]{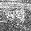} &
		\includegraphics[width = 0.1625\textwidth]{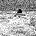} &
		\includegraphics[width = 0.1625\textwidth]{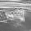} &
		\includegraphics[width = 0.1625\textwidth]{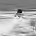} \\
		
		\multicolumn{2}{c}{Ground truth}&
		\multicolumn{2}{c}{Noisy image}&
		\multicolumn{2}{c}{BM3D}\\
		
		\multicolumn{2}{c}{}&
		\multicolumn{2}{c}{}&
		\multicolumn{2}{c}{$27.59dB$}\\
		
		\smallskip
		
	\end{tabular} 
	\begin{tabular}{c@{\hskip 0.005\textwidth}c@{\hskip 0.005\textwidth}c}
		
		\includegraphics[width = 0.33\textwidth]{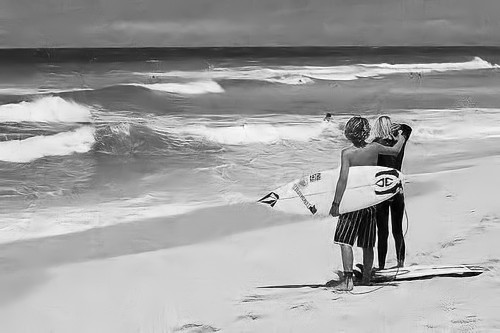} &
		\includegraphics[width = 0.33\textwidth]{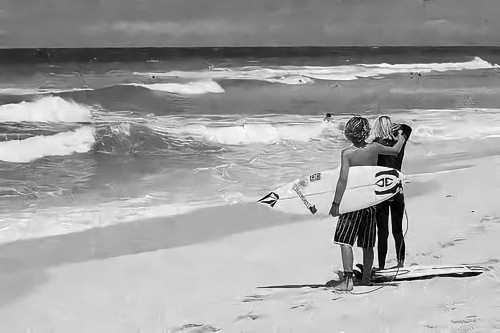} &
		\includegraphics[width = 0.33\textwidth]{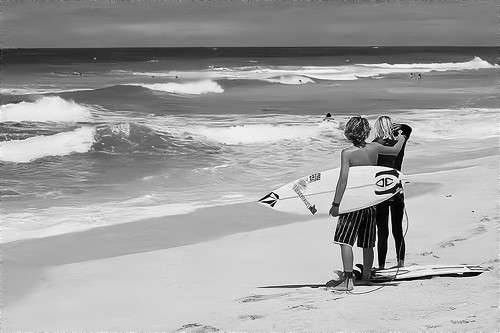} \\
		
	\end{tabular}  
	\begin{tabular}{c@{\hskip 0.005\textwidth}c@{\hskip 0.005\textwidth}c@{\hskip 0.005\textwidth}c@{\hskip 0.005\textwidth}c@{\hskip 0.005\textwidth}c}
		
		\includegraphics[width = 0.1625\textwidth]{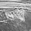} &
		\includegraphics[width = 0.1625\textwidth]{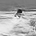} &
		\includegraphics[width = 0.1625\textwidth]{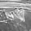} &
		\includegraphics[width = 0.1625\textwidth]{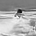} &
		\includegraphics[width = 0.1625\textwidth]{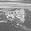} &
		\includegraphics[width = 0.1625\textwidth]{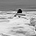} \\
		
		\multicolumn{2}{c}{MLP}&
		\multicolumn{2}{c}{TNRD}&
		\multicolumn{2}{c}{\textit{Beach}-specific DenoiseNet}\\
		
		\multicolumn{2}{c}{$27.84dB$}&
		\multicolumn{2}{c}{$27.81dB$}&
		\multicolumn{2}{c}{$28.40dB$}\\
		
		\smallskip
		
	\end{tabular}   
\end{figure*}
\bigskip 

\begin{figure*}[]  
	\centering     
	\begin{tabular}{c@{\hskip 0.005\textwidth}c@{\hskip 0.005\textwidth}c}
		
		\includegraphics[width = 0.33\textwidth]{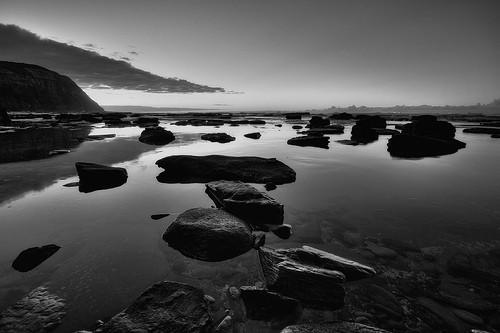} &
		\includegraphics[width = 0.33\textwidth]{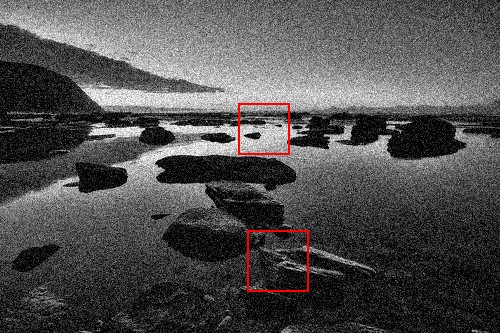} &
		\includegraphics[width = 0.33\textwidth]{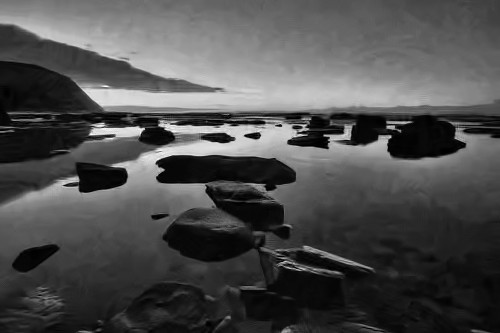} \\
		
	\end{tabular}  
	\begin{tabular}{c@{\hskip 0.005\textwidth}c@{\hskip 0.005\textwidth}c@{\hskip 0.005\textwidth}c@{\hskip 0.005\textwidth}c@{\hskip 0.005\textwidth}c}
		
		\includegraphics[width = 0.1625\textwidth]{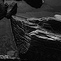} &
		\includegraphics[width = 0.1625\textwidth]{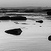} &
		\includegraphics[width = 0.1625\textwidth]{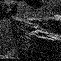} &
		\includegraphics[width = 0.1625\textwidth]{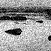} &
		\includegraphics[width = 0.1625\textwidth]{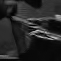} &
		\includegraphics[width = 0.1625\textwidth]{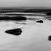} \\
		
		\multicolumn{2}{c}{Ground truth}&
		\multicolumn{2}{c}{Noisy image}&
		\multicolumn{2}{c}{BM3D}\\
		
		\multicolumn{2}{c}{}&
		\multicolumn{2}{c}{}&
		\multicolumn{2}{c}{$30.85dB$}\\
		
		\smallskip        
	\end{tabular} 
	\begin{tabular}{c@{\hskip 0.005\textwidth}c@{\hskip 0.005\textwidth}c}
		
		\includegraphics[width = 0.33\textwidth]{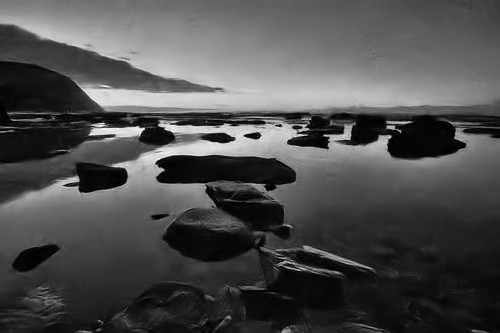} &
		\includegraphics[width = 0.33\textwidth]{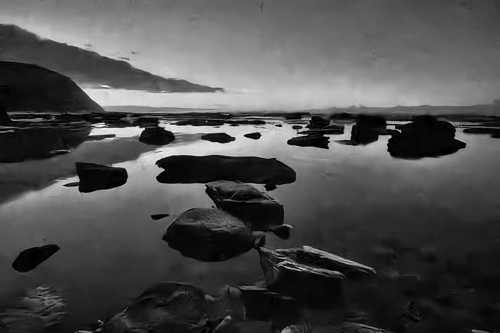} &
		\includegraphics[width = 0.33\textwidth]{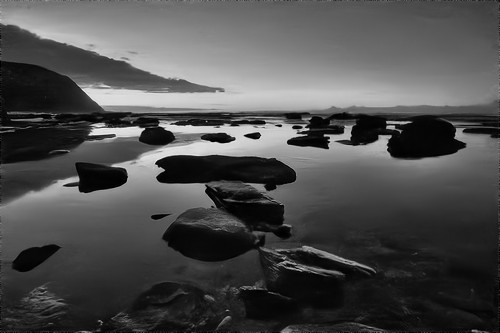} \\
		
	\end{tabular}  
	\begin{tabular}{c@{\hskip 0.005\textwidth}c@{\hskip 0.005\textwidth}c@{\hskip 0.005\textwidth}c@{\hskip 0.005\textwidth}c@{\hskip 0.005\textwidth}c}
		
		\includegraphics[width = 0.1625\textwidth]{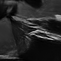} &
		\includegraphics[width = 0.1625\textwidth]{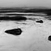} &
		\includegraphics[width = 0.1625\textwidth]{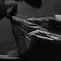} &
		\includegraphics[width = 0.1625\textwidth]{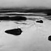} &
		\includegraphics[width = 0.1625\textwidth]{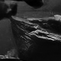} &
		\includegraphics[width = 0.1625\textwidth]{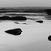} \\
		
		\multicolumn{2}{c}{MLP}&
		\multicolumn{2}{c}{TNRD}&
		\multicolumn{2}{c}{\textit{Beach}-specific DenoiseNet}\\
		
		\multicolumn{2}{c}{$31.25dB$}&
		\multicolumn{2}{c}{$31.30dB$}&
		\multicolumn{2}{c}{$31.87dB$}\\
		
		\smallskip        
	\end{tabular} 
\end{figure*}
\bigskip

\end{document}